\documentclass[10pt,twocolumn,letterpaper]{article}

\usepackage[pagenumbers]{cvpr}

\usepackage[dvipsnames]{xcolor}
\usepackage{booktabs}
\usepackage{graphicx}
\usepackage{multirow}
\usepackage{placeins}
\usepackage{stfloats}
\usepackage{adjustbox}
\usepackage{threeparttable}
\usepackage[utf8]{inputenc}
\DeclareUnicodeCharacter{03B2}{\ensuremath{\beta}}
\usepackage{capt-of}
\usepackage[accsupp]{axessibility}

\newcommand{\ADNIAckText}{%
Data collection and sharing for the Alzheimer's Disease Neuroimaging Initiative (ADNI) is funded by the National Institute on Aging (National Institutes of Health Grant U19 AG024904). The grantee organization is the Northern California Institute for Research and Education. In the past, ADNI has also received funding from the National Institute of Biomedical Imaging and Bioengineering, the Canadian Institutes of Health Research, and private sector contributions through the Foundation for the National Institutes of Health (FNIH) including generous contributions from the following: AbbVie, Alzheimer’s Association; Alzheimer’s Drug Discovery Foundation; Araclon Biotech; BioClinica, Inc.; Biogen; Bristol-Myers Squibb Company; CereSpir, Inc.; Cogstate; Eisai Inc.; Elan Pharmaceuticals, Inc.; Eli Lilly and Company; EuroImmun; F. Hoffmann-La Roche Ltd and its affiliated company Genentech, Inc.; Fujirebio; GE Healthcare; IXICO Ltd.; Janssen Alzheimer Immunotherapy Research \& Development, LLC.; Johnson \& Johnson Pharmaceutical Research \&Development LLC.; Lumosity; Lundbeck; Merck \& Co., Inc.; Meso Scale Diagnostics, LLC.; NeuroRx Research; Neurotrack Technologies; Novartis Pharmaceuticals Corporation; Pfizer Inc.; Piramal Imaging; Servier; Takeda Pharmaceutical Company; and Transition Therapeutics.}

\newcommand{\OASISAckText}{%
Data were provided in part by OASIS\nobreakdash-3 (Longitudinal Multimodal Neuroimaging; Principal Investigators: T.~Benzinger, D.~Marcus, J.~Morris; NIH P30~AG066444, P50~AG00561, P30~NS09857781, P01~AG026276, P01~AG003991, R01~AG043434, UL1~TR000448, R01~EB009352). AV\nobreakdash-45 doses were provided by Avid Radiopharmaceuticals, a wholly owned subsidiary of Eli Lilly.}

\definecolor{cvprblue}{rgb}{0.21,0.49,0.74}
\usepackage[pagebackref,breaklinks,colorlinks,allcolors=cvprblue]{hyperref}
\usepackage{xurl}

\def\paperID{16}
\def\confName{CVPR}
\def\confYear{2026}

\title{Cross-Modal Knowledge Distillation for PET-Free\\
Amyloid\nobreakdash-Beta Detection~from~MRI}

\author{
Francesco Chiumento$^{1}$ \quad
Julia Dietlmeier$^{1,2}$ \quad
Ronan P. Killeen$^{3,4}$\\
Kathleen M. Curran$^{2,4}$ \quad
Noel E. O'Connor$^{1,2}$ \quad
Mingming Liu$^{1,2}$\\
$^{1}$Dublin City University, Ireland \quad
$^{2}$Insight Research Ireland Centre for Data Analytics, Ireland\\
$^{3}$St. Vincent's University Hospital, Ireland \quad
$^{4}$University College Dublin, Ireland\\
{\tt\small francesco.chiumento2@mail.dcu.ie}
}

\begin{document}

\captionsetup{hypcap=false}

\twocolumn[{
  \maketitle
  
  \begin{center}
    \vspace{-2em}
    
    \includegraphics[width=0.95\textwidth]{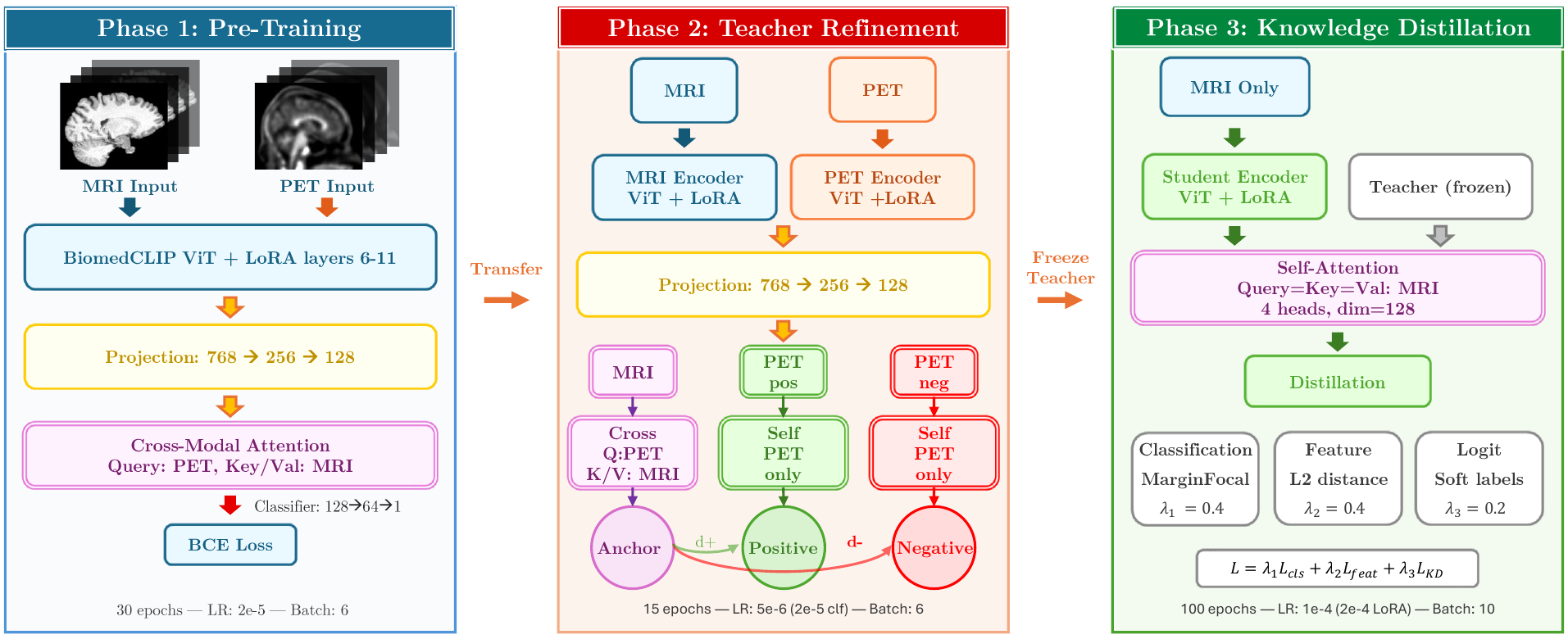}
    
    \includegraphics[width=0.57\textwidth]{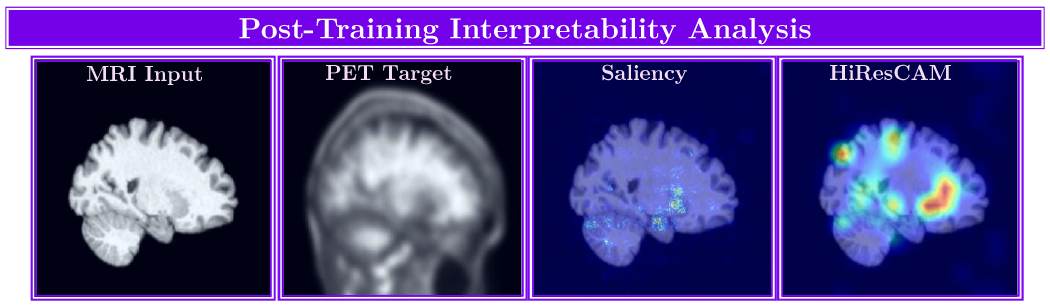}
    
    \captionof{figure}{Overview of our PET-guided knowledge distillation framework. 
    A teacher model learns cross-modal alignment between PET and MRI, guided by Centiloid-aware triplet mining. The teacher's knowledge is then distilled to an MRI-only student for PET-free inference.}
    \label{fig:teaser}

  \end{center}
}]

\begin{abstract}
Detecting amyloid-$\beta$ (A$\beta$) positivity is crucial for early diagnosis of Alzheimer's disease but typically requires PET imaging, which is costly, invasive, and not widely accessible, limiting its use for population-level screening. We address this gap by proposing a PET-guided knowledge distillation framework that enables A$\beta$ prediction from MRI alone, without requiring non-imaging clinical covariates or PET at inference. Our approach employs a BiomedCLIP-based teacher model that learns PET--MRI alignment via cross-modal attention and triplet contrastive learning with PET-informed (Centiloid-aware) online negative sampling. An MRI-only student then mimics the teacher via feature-level and logit-level distillation. Evaluated across four MRI contrasts (T1w, T2w, FLAIR, T2*) and two independent datasets, our approach demonstrates effective knowledge transfer (best AUC: 0.74 on OASIS-3, 0.68 on ADNI) while maintaining interpretability and eliminating the need for clinical variables. Saliency analysis confirms that predictions focus on anatomically relevant cortical regions, supporting the clinical viability of PET-free A$\beta$ screening. Code is available at \href{https://github.com/FrancescoChiumento/pet-guided-mri-amyloid-detection}{github.com/FrancescoChiumento/pet-guided-mri-amyloid-detection}.
\end{abstract}
\section{Introduction}

Alzheimer’s disease (AD) is a progressive neurodegenerative disorder that represents the predominant form of dementia. 55.2 million individuals are currently affected worldwide, with projections reaching 78 million by 2030 and costs projected to reach up to \$2.8~trillion~\cite{whoBlueprintDementiaResearch2022,monteiroAlzheimersDiseaseInsights2023a}.
The pathological hallmarks of AD are neurofibrillary tangles (NFTs), formed by hyperphosphorylated tau protein within neurons, and extracellular plaques of accumulated A$\beta$ peptide accompanied by neuroinflammation, synaptic dysfunction, mitochondrial and vascular abnormalities~\cite{zhangRecentAdvancesAlzheimers2024,safiriAlzheimersDiseaseComprehensive2024}.
Currently, there is no effective cure for AD, and once it manifests, it follows an irreversible progression~\cite{zhangRecentAdvancesAlzheimers2024,gunesBiomarkersAlzheimersDisease2022}. 

An early event in AD pathophysiology is the deposition of A$\beta$ plaques in the brain, which can occur ${\sim}20$~years before the onset of symptoms, offering a potential window for secondary prevention~\cite{caselliNeuropsychologicalDecline202020}. Typically, A$\beta$ is assessed through invasive procedures such as positron emission tomography (PET) or cerebrospinal fluid (CSF) assays. PET uses radiotracers targeting amyloid or tau pathology, creating a spatial map of A$\beta$ distribution in the brain and revealing the extent of AD pathology~\cite{baoPETNeuroimagingAlzheimers2021}. Although amyloid levels can be measured in individuals using PET imaging with amyloid-sensitive ligands such as Pittsburgh Compound B (PiB; $^{11}\mathrm{C}$-PiB) or florbetapir ($^{18}\mathrm{F}$-AV-45)~\cite{yaminPittsburghCompoundBPiB2017,camusUsingPET18FAV452012a}, amyloid-PET is expensive, not widely available, and invasive, exposing the patient to ionizing radiation. For this reason, amyloid-PET is not considered cost-effective for diagnostic use at the MCI stage~\cite{leeCosteffectivenessUsingAmyloid2021}, yet demand is projected to increase up to ${\sim}$20-fold following approval of anti-A$\beta$ therapies (lecanemab, donanemab) that require confirmation of amyloid pathology prior to treatment initiation~\cite{vergerFDAApprovalLecanemab2023a,u.s.foodanddrugadministrationLeqembiLecanemabirmbInjection2023a,u.s.foodanddrugadministrationKISUNLADonanemabazbtInjection2024a}.

MRI, conversely, is non-invasive, widely accessible, and provides detailed structural brain information without radioactive tracers~\cite{wangDeeplearningBasedMultimodal2025}. While MRI cannot directly visualize amyloid, A$\beta$ accumulation leads to widespread brain cell loss, which manifests as atrophy in T1-weighted (T1w) sequences~\cite{chattopadhyayComparisonDeepLearning2024a}.
Several studies classify A$\beta$ positivity (A$\beta$+ vs.\ A$\beta$-) from PET images using deep learning models. A smaller subset uses MRI, but typically includes additional non-imaging covariates such as APOE$\varepsilon$4 genotype, cognitive scores, hippocampal volumetry, or clinical diagnosis~\cite{dolciMultimodalMRIAccurately2025,kimPredictingAmyloidPositivity2021,lewMRIbasedDeepLearning2023,wangDeeplearningBasedMultimodal2025}; however, these covariates exhibit high 
missingness in practice: routine APOE testing is not recommended by ACMG guidelines~\cite{goldmanGeneticCounselingTesting2011a} and is missing in 26\% of NACC participants \cite{pirragliaSubtypesMultipleetiologyDementias2025b}; cognitive measures are absent in 89\% of EHR dementia records~\cite{maserejianCognitiveMeasuresLacking2021b}; and only 23\% of European centres perform volumetry analysis~\cite{vernooijDementiaImagingClinical2019b}.
Few studies, on the other hand, have attempted to predict amyloid status using only MRI data without other clinical variables~\cite{kimDeepLearningBasedPrediction2025}.

In this work, we address binary A$\beta$+ prediction using only MRI as the single modality. Specifically, our research question is how to effectively transfer the knowledge from a multimodal contrastive PET and MRI \emph{teacher} to an MRI-only \emph{student}. With this in mind, the key novelty of our work is that we propose a PET-guided \emph{knowledge distillation} framework based on BiomedCLIP, a ViT pre-trained on 15M biomedical image-text pairs~\cite{zhangBiomedCLIPMultimodalBiomedical2025a} which, to the best of our knowledge, has not been applied in this specific context. The cross-modal \emph{teacher} integrates PET (query) and MRI (key) via multi-head attention (MHA) and is supervised by a combination of classification and \emph{triplet loss}; the MRI-only \emph{student} learns from the \emph{teacher} through feature-level and logit-level distillation with temperature annealing. At inference time, only MRI is required.
Fig.~\ref{fig:teaser} illustrates our three-phase distillation approach. The main contributions of our work are outlined below.
\begin{itemize}
    \item \textbf{PET-to-MRI knowledge distillation framework:} We propose a novel cross-modal distillation approach for amyloid prediction, where an MRI-only student learns from a PET+MRI teacher via $\ell_2$-normalized feature matching and temperature-scaled logit distillation, enabling PET-free inference without non-imaging covariates. Unlike prior work using direct supervision from binary PET labels or CSF values, our teacher learns from PET's spatial standardized uptake value ratio (SUVR) distribution~\cite{klunkCentiloidProjectStandardizing2015} through cross-modal attention, transferring spatially-informed knowledge to the student model.
    
    \item \textbf{Contrastive teacher training:} 
    We propose a new adaptation of the triplet loss with Centiloid (CL, a standardized PET amyloid quantification scale~\cite{klunkCentiloidProjectStandardizing2015,iaccarinoPracticalOverviewUse2025})\nobreakdash-aware negative mining, where online hard-negative selection ensures biochemically distinct triplets, while cross-modal attention and self-attention synthesize patient-level embeddings for triplet learning.
    
    \item \textbf{Multi-contrast evaluation and cross-dataset transfer:} 
    We evaluate T1w, T2w, FLAIR, and T2* sequences and their combinations on OASIS-3 and ADNI, achieving AUCs up to 0.74 and improving T1w MRI-only results by +19.7\% (0.73 vs.\ 0.61) over Kim et al.~\cite{kimDeepLearningBasedPrediction2025}, and validate cross-dataset knowledge distillation (CDKD)~\cite{alzheimersdiseaseneuroimaginginitiativeadniAlzheimersDiseaseNeuroimaging2025,lamontagneOASIS3LongitudinalNeuroimaging2019d}.
    
\end{itemize}

\section{Related Work}
\label{sec:related_works}

\subsection{MRI-based Amyloid Prediction}
MRI-only methods for predicting A$\beta$+ remain limited and achieve only moderate discriminative performance. 
Most studies focus on binary classification of amyloid-$\beta$ status (A$\beta$+ vs.\ A$\beta$-) in AD research, often using PET or CSF labels within deep learning models~\cite{dolciMultimodalMRIAccurately2025}. A smaller number of works have investigated MRI-only prediction. 
Kim et al.~\cite{kimDeepLearningBasedPrediction2025} trained a 3D EfficientNet on 4{,}056 exams (ADNI, OASIS-3, and A4), reporting AUCs of 0.61 for T1w and 0.67 for T1w+FLAIR, without clinical variables. In contrast, when structured variables are added, performance increases: Lew et al.~\cite{lewMRIbasedDeepLearning2023} combined MRI with demographic, APOE, and cognitive information to predict PET-derived Amyloid/Tau/Neurodegeneration (ATN) status, obtaining an amyloid AUC of 0.79 using SUVR thresholded via Gaussian mixture modeling.
Most prior work either relies on direct supervision or clinical variables, with few truly MRI-only pipelines and no exploitation of PET--MRI complementarity during training~\cite{kimDeepLearningBasedPrediction2025}. We address this by supervising MRI features with a PET-guided teacher through cross-modal distillation.

Although structured covariates can improve prediction, their use reduces deployability in practice, since key variables (e.g., APOE genotype and detailed cognitive scores) are not consistently available across imaging workflows and cohorts~\cite{ritchieApolipoproteinGeneticTesting2024, renMovingMedicalStatistics2024, pirragliaSubtypesMultipleetiologyDementias2025b, maserejianCognitiveMeasuresLacking2021b, vernooijDementiaImagingClinical2019b}. Also, beyond T1w, integration of additional contrasts such as FLAIR or DTI has been explored in a few studies, but evidence for T2w or T2*-GRE/SWI sequences for A$\beta$ classification remains limited, as these modalities are more commonly used for amyloid-related imaging abnormalities (ARIA) or vascular assessment~\cite{roytmanAmyloidRelatedImagingAbnormalities2023}. Conversely, MRI scans are routinely acquired and recommended as \emph{first-line} neuroimaging, making them a pragmatic and scalable input for automated screening models~\cite{rajiValueNeuroimagingDementia2022}. This supports multimodal supervision with single-modality inference.

\subsection{Multimodal Fusion and Its Limitations}
Multimodal approaches (structural, functional, and diffusion MRI) have shown promising results~\cite{dolciMultimodalMRIAccurately2025} but require all modalities to be available at both training and inference time, limiting real-world applicability. These limitations have motivated alternative strategies that exploit multimodal information during training while enabling single-modality inference. Among these, \emph{knowledge distillation} has emerged as a framework for cross-modal transfer~\cite{liKnowledgeDistillationTeacher2026}.

\subsection{Knowledge Distillation in Medical Imaging}
Knowledge distillation (KD) is a learning paradigm in which a compact ``student'' network learns from a high-capacity ``teacher'' network, improving efficiency or enabling cross-modal transfer.
In medical imaging, KD has been applied to tasks such as segmentation, classification, and reconstruction. To the best of our knowledge, PET-to-MRI distillation for amyloid prediction has not been extensively explored~\cite{liKnowledgeDistillationTeacher2026}. Building on this overview, we identify key limitations in the current literature:

\begin{itemize}
    \item No use of PET--MRI cross-modal supervision during training;
    \item Limited evaluation of multi-contrast MRI beyond T1w+FLAIR;
    \item Limited assessment of model generalization across independent cohorts.
\end{itemize}

We adopt CL to define amyloid status because CL provides a tracer- and cohort-independent scale~\cite{kumarAnalysingHeterogeneityAlzheimers2024, iaccarinoPracticalOverviewUse2025}, enabling cross-dataset transfer between OASIS-3 and ADNI. In contrast, some prior work uses SUVR-based thresholding (including data-driven Gaussian mixture cutoffs)~\cite{lewMRIbasedDeepLearning2023} or CSF-based definitions~\cite{dolciMultimodalMRIAccurately2025}, which are internally consistent but less comparable across sites, tracers and cohorts~\cite{iaccarinoPracticalOverviewUse2025}. Since label definitions and targets differ, absolute AUCs are indicative rather than strictly interchangeable.
\section{Datasets and Cohorts}

\paragraph{Public cohorts}
We evaluate two large public cohorts for aging and AD (Table~\ref{tab:publiccohorts}), with subsets detailed in Table~\ref{tab:cohorts_compact}.

\begin{itemize}
    \item \textbf{OASIS-3}~\cite{lamontagneOASIS3LongitudinalNeuroimaging2019d} is a longitudinal dataset spanning $\sim$30 years, with 1{,}379 subjects, 2{,}842 MRI sessions, and 2{,}157 PET sessions. PET data include SUVR and CL quantification via the PET Unified Pipeline (PUP)~\cite{suYsu001PUP2025}.
    
    \item \textbf{ADNI}~\cite{alzheimersdiseaseneuroimaginginitiativeadniAlzheimersDiseaseNeuroimaging2025} is a widely used multicenter longitudinal study with over 5{,}100 participants (3{,}262 with MRI, 1{,}089 with PET) sharing imaging and structured data under the ADNI Data Use Agreement and publication policy.
\end{itemize}

\begin{table}[!htb]
\centering
\caption{\textbf{Public cohorts overview.} Amyloid PET tracers used.}
\label{tab:publiccohorts}
\begin{adjustbox}{max width=0.90\columnwidth}
\begin{tabular}{@{}llll@{}}
\toprule
\textbf{Cohort} & \textbf{Subjects} & \textbf{PET Tracers} & \textbf{Notes} \\
\midrule
OASIS-3~\cite{lamontagneOASIS3LongitudinalNeuroimaging2019d} & 1{,}379 & PiB, AV\nobreakdash-45 & Multimodal; PUP \\
ADNI~\cite{alzheimersdiseaseneuroimaginginitiativeadniAlzheimersDiseaseNeuroimaging2025} & 5{,}111 & PiB, AV\nobreakdash-45 & Multicenter \\
\bottomrule
\end{tabular}
\end{adjustbox}
\end{table}

\paragraph{Data usage statement}
Data are obtained from the longitudinal databases OASIS-3~\cite{lamontagneOASIS3LongitudinalNeuroimaging2019d} and the Alzheimer's Disease Neuroimaging Initiative (ADNI)~\cite{alzheimersdiseaseneuroimaginginitiativeadniAlzheimersDiseaseNeuroimaging2025}. We construct modality-consistent MRI cohorts using the same procedure across both datasets:
\begin{itemize}
    \item (\textit{T1w}~$\cap$~\textit{T2w}): all MRI sessions where both T1w and T2w are available for the same subject/timepoint;
    \item (\textit{T2*}~$\cap$~\textit{FLAIR}): all MRI sessions where both T2* and FLAIR are available for the same subject/timepoint.
\end{itemize}

Each cohort is partitioned with identical modality constraints; sessions missing contrasts are excluded (no imputation). We retain all timepoints matching the modality combination, considering only subjects with MRI and PET acquired within 365~days. Data splits are performed at the subject level to avoid subject overlap and prevent data leakage across folds. When a subject has multiple PET tracers (PiB, AV\nobreakdash-45), all scans are assigned to the same fold.

\begin{table}[!ht]
\centering
\caption{\textbf{Cohorts used.} Splits are subject-wise.}
\label{tab:cohorts_compact}
\setlength{\tabcolsep}{3.2pt} 
\scriptsize                  
\begin{tabular}{@{}lccc ccc c@{}}
\toprule
\textbf{Cohort (MRI)} &
\multicolumn{3}{c}{\textbf{Sessions}} &
\multicolumn{3}{c}{\textbf{Subjects}} &
\textbf{\%Pos (Train)} \\
\cmidrule(lr){2-4}\cmidrule(lr){5-7}
& \textbf{Train} & \textbf{Val} & \textbf{Test}
& \textbf{Train} & \textbf{Val} & \textbf{Test}
& \\
\midrule
OASIS-3 (T1w~$\cap$~T2w)   & 518 & 187 & 166 & 346 & 118 & 127 & 21.8 \\
OASIS-3 (FLAIR~$\cap$~T2*) & 288 & 96 & 89 & 287 & 93 & 88 & 34.4 \\
ADNI (T1w~$\cap$~T2w)      & 303 & 102 &  90 & 212 &  72 &  68 & 44.9 \\
ADNI (FLAIR~$\cap$~T2*)    & 608 & 193 & 190 & 405 & 132 & 132 & 50.2 \\
\bottomrule
\end{tabular}
\end{table}  
\section{Methodology}

\subsection{Overview and Framework Architecture}
\label{sec:overview}

Our methodology addresses A$\beta$ detection without PET images through a three-phase distillation framework (Figs.~\ref{fig:phase1-pretrain}--\ref{fig:phase3-kd}).
We train a teacher that captures shared PET--MRI patterns, then distill this knowledge into a student operating on MRI-only (single- or multi-contrast: T1-weighted (T1w), T2-weighted (T2w), fluid-attenuated inversion recovery (FLAIR), and T2*-weighted (T2*)). The framework consists of:
\begin{enumerate}
    \item \textbf{Reference labels}: binary A$\beta$ status from PET-derived CL scores (positive if CL $> 20.6$).
    \item \textbf{Preprocessing and registration}: N4, HD-BET, ANTs registration to MNI, and slice-level PET--MRI pairing.
    \item \textbf{Architecture}: BiomedCLIP ViT with LoRA; cross-modal attention (teacher), self-attention (student).
    \item \textbf{Training}: Phase~1 pre-training with classification (Sec.~\ref{sec:phase1}), Phase~2 contrastive learning with \emph{CL-aware} online negative sampling for triplets (selecting negatives with $|\text{CL}_{\text{anchor}} - \text{CL}_{\text{neg}}| \geq 5.0$) (Sec.~\ref{sec:phase2}), and Phase~3 feature and logit distillation (Sec.~\ref{sec:phase3}, Figs.~\ref{fig:phase1-pretrain}--\ref{fig:phase3-kd}).

\end{enumerate}

\noindent At test time, only MRI is required; PET images and CL values are not needed.

\subsection{Ground Truth: PET-Based Amyloid Quantification}
\label{sec:ground_truth}

We use PET-derived Centiloid (CL) values as the ground truth for binary amyloid status, which provide a tracer-harmonized scale across AV\nobreakdash-45 and PiB acquisitions \cite{iaccarinoPracticalOverviewUse2025,klunkCentiloidProjectStandardizing2015}. OASIS-3 uses the PET Unified Pipeline (PUP) \cite{suYsu001PUP2025} for cortical SUVR normalized to cerebellum. ADNI uses Berkeley's \texttt{CENTILOIDS} field (\texttt{UCBERKELEY\_AMY\_6MM}) for AV\nobreakdash-45 quantification~\cite{royseValidationAmyloidPET2021}. We apply a fixed threshold \textbf{CL $> 20.6$} uniformly across cohorts consistent with established practice~\cite{amadoruComparisonAmyloidPET2020,OASIS_3_Imaging_Methods,lehmannComparativePerformancePlasma2025,klunkCentiloidProjectStandardizing2015}, without demographic or clinical variables.

The use of CL values serves two distinct purposes: \textbf{(i)~training}---binary label and \emph{CL-aware negative sampling} in triplet mining (teacher accesses PET spatial distribution); \textbf{(ii)~inference}---MRI-only, model outputs $p(\text{A}\beta+)$ thresholded at $\theta = 0.5$ (fixed) or $\theta^*$ (validation-optimized).
\noindent Table~\ref{tab:cohorts_compact} shows the resulting A$\beta$ prevalence across cohorts; OASIS-3 exhibits class imbalance.

\subsection{Data Organization and Preprocessing}
\label{sec:preproc}
\paragraph{MRI Preprocessing}
MRI and PET scans are temporally paired within $|\Delta t_{\text{MRI--PET}}| \leq 365$ days, consistent with low annualized CL change \cite{bollackInvestigatingReliableAmyloid2024a}. 
For each MRI contrast (T1w, T2w, FLAIR, T2*), we apply:  
(i) N4 bias field correction with Otsu thresholding using 200 histogram bins to generate a binary foreground mask, excluding background voxels from correction \cite{tustisonN4ITKImprovedN32010};  
(ii) HD-BET brain extraction~\cite{isenseeAutomatedBrainExtraction2019a}, a deep learning-based method trained on multi-site data. HD-BET skull stripping (mask + brain volume) employs test-time augmentation for improved robustness;  
(iii) robust intensity normalization: clip 0.5--99.5\%, min--max to $[0,1]$, then gamma correction $I_{\text{norm}} = I^{0.9}$ for gray--white matter contrast \cite{chiumentoDetectingBetaAmyloidCrossModal2025}.

\paragraph{PET--MRI Registration}
PET volumes are smoothed with a Gaussian kernel (FWHM 8~mm) following the PET Unified Pipeline protocol~\cite{suYsu001PUP2025} and rigidly aligned to their corresponding skull-stripped MRI using ANTs (\texttt{type\_of\_transform='Rigid'}) \cite{avantsSymmetricDiffeomorphicImage2008}, then both are registered to the MNI ICBM152 symmetric template using rigid transformation only~\cite{fonovUnbiasedAverageAgeappropriate2011,fonovUnbiasedNonlinearAverage2009}. Finally, the MRI-to-MNI transform is applied to the PET volume previously aligned to MRI via \texttt{ants.apply\_transforms} with linear interpolation, yielding standardized files. We use \emph{only rigid transformations} (6 degrees of freedom: 3 translations + 3 rotations) to avoid introducing artificial deformations that could misrepresent amyloid distribution patterns.

\paragraph{Slice Extraction and Pairing}
After MNI registration ($1{\times}1{\times}1$~mm$^3$ isotropic), from each volume we extract $S_{\text{full}} = 40$ sagittal slices (resized to $224{\times}224$ for ViT input) uniformly from the central 40--60\% along the left--right axis. During training, we subsample $S_{\text{train}} = 25$ uniformly spaced slices balancing coverage, batch diversity, and efficiency, leveraging BiomedCLIP 2D pretraining with multi-head attention pooling. PET--MRI pairs use matching slice indices for anatomical correspondence. Slices with mean intensity $< 0.01$ (background) are discarded. We create one fixed subject-level split using \texttt{StratifiedGroupKFold} (K=5, seed=42): 3 folds train, 1 val, 1 test; all sessions/tracers per subject remain within one fold to prevent leakage. Stratification is based on the binary amyloid label to maintain class balance. We also check tracer proportions across splits to avoid distribution drift.

\paragraph{Training Strategies}
During Phases 1--2, we apply synchronized spatial and intensity augmentations (affine transforms, color jitter, blur/noise, random erasing) with shared random seeds for PET--MRI pairs to preserve anatomical correspondence (see supplementary). To address class imbalance (Table~\ref{tab:cohorts_compact}), we use \texttt{WeightedRandomSampler} with inverse frequency weights ($w_i = 1/n_{y_i}$), ensuring balanced mini-batches.

\subsection{Network Architecture}
\label{sec:architecture}

We adapt a BiomedCLIP Vision Transformer (ViT) encoder \cite{zhangBiomedCLIPMultimodalBiomedical2025a} using Low-Rank Adaptation (LoRA) \cite{huLoRALowRankAdaptation2021c} applied to the last six transformer blocks (layers 6--11, 0-indexed; rank $r = 32$) on attention projections (\textit{q/k/v/out}), while keeping earlier blocks frozen. LoRA introduces trainable low-rank matrices that reparameterize frozen pretrained weights, enabling efficient fine-tuning (the formula and initialization can be found in the supplementary).

Each modality (MRI or PET) is processed as a stack of $S$ slices resized to $224{\times}224$. Per-slice CLS tokens (768D) are projected to 128D through a two-layer nonlinear head with LayerNorm, GELU, and dropout. Slice embeddings are then processed by an MHA layer with 4 heads and pooled via learned attention weights $\alpha_s = \mathrm{softmax}(w_s/\tau)$ with temperature $\tau{=}2.0$ to obtain patient-level representations $\mathbf{e} = \sum_s \alpha_s\,\mathbf{A}_s$ (see supplementary).

\paragraph{Attention Mechanisms}
\textbf{Teacher:} extracts MRI patterns predictive of PET-derived amyloid pathology through cross-modal attention:
\begin{equation}
\mathbf{A}^{(T)} = \mathrm{CrossAttn}(\mathbf{Q} = \mathbf{E}_{\text{PET}},\,\mathbf{K} = \mathbf{V} = \mathbf{E}_{\text{MRI}}),
\end{equation}
where $\mathbf{E}_{\text{PET}}, \mathbf{E}_{\text{MRI}} \in \mathbb{R}^{B \times S \times 128}$ are batched slice embeddings, then attention-pooled to obtain $\mathbf{e}_T = \sum_{s}\alpha^{(T)}_s\,\mathbf{A}^{(T)}_s$.
\noindent\textbf{Student:} aggregates MRI-only features via self-attention:
\begin{equation}
\mathbf{A}^{(S)} = \mathrm{SelfAttn}(\mathbf{Q} = \mathbf{K} = \mathbf{V} = \mathbf{E}_{\text{MRI}}),
\end{equation}
followed by the same pooling to produce $\mathbf{e}_S = \sum_{s}\alpha^{(S)}_s\,\mathbf{A}^{(S)}_s$. We match $\ell_2$-normalized embeddings $\hat{\mathbf{e}}_S$ to teacher embeddings $\hat{\mathbf{e}}_T$ via feature distillation (Sec.~\ref{sec:distillation-loss}).

\paragraph{Classifier}
Both models use a two-layer MLP with ReLU and dropout ($128 \to 64 \to 1$) to produce the logit $z$. Dropout rates are higher during teacher training (Phases 1--2) and reduced during distillation.

\subsection{Phase 1: Teacher Pre-Training}
\label{sec:phase1}

\begin{figure}[h!t]
  \centering
  \includegraphics[width=1.00\columnwidth]{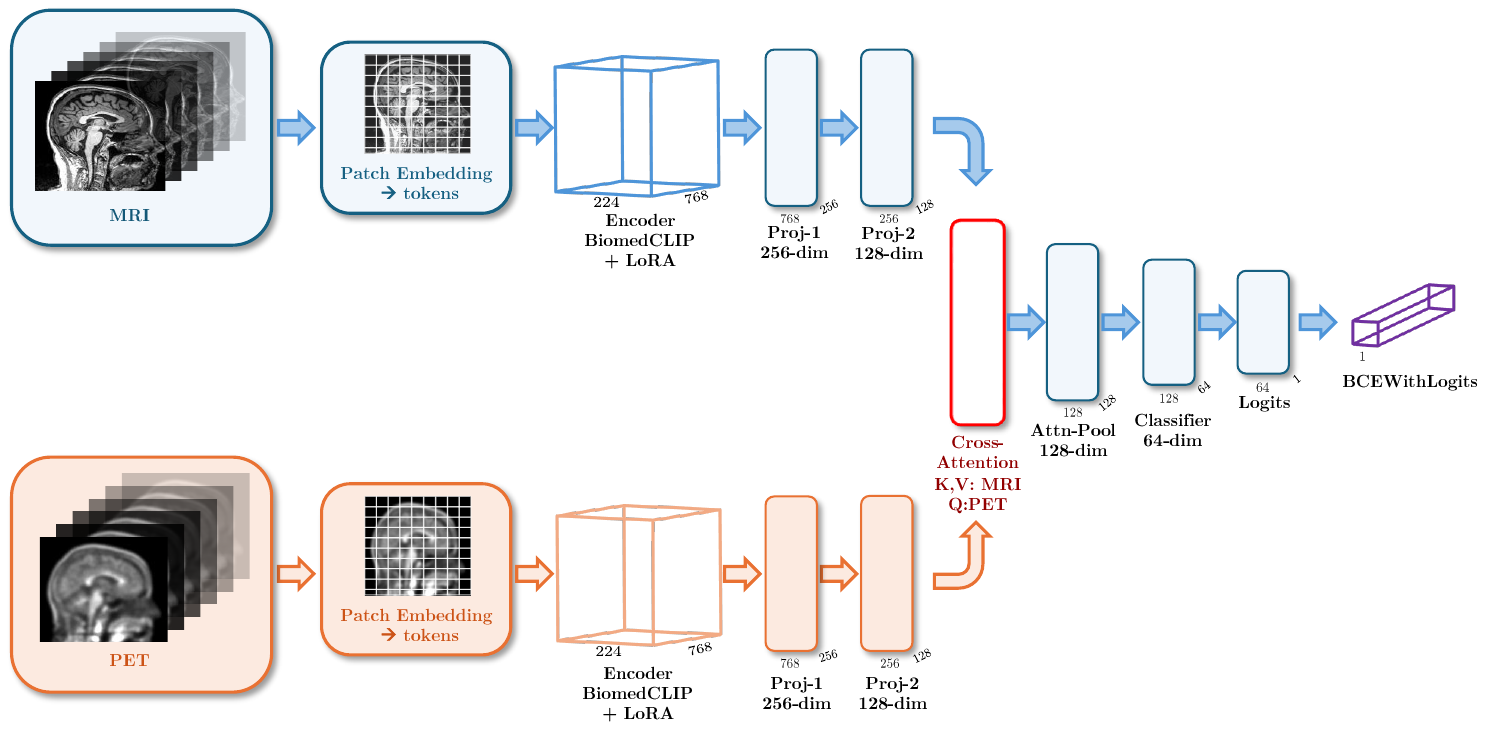}
   \caption{\textbf{Phase 1 --- Teacher pre-training} with binary classification loss (BCE).}
  \label{fig:phase1-pretrain}
\end{figure}

We pre-train the teacher on PET$\rightarrow$MRI embeddings (Fig.~\ref{fig:phase1-pretrain}): $\mathbf{e}_T = \mathrm{AttnPool}(\mathbf{E}_{\text{PET}},\mathbf{E}_{\text{MRI}})$, $z_T=\mathrm{MLP}(\mathbf{e}_T)$, using AdamW \cite{loshchilovDecoupledWeightDecay2019} (lr $2 \times 10^{-5}$, weight decay (wd) $10^{-3}$, batch size 6) with binary cross-entropy with logits (BCE) for 30 epochs with gradient clipping and warm-up (see supplementary).

\subsection{Phase 2: Contrastive Teacher Refinement}
\label{sec:phase2}

\paragraph{CL-Aware Online Negative Sampling}

We perform \emph{online} hard-negative mining using PET-derived CL values during training. Triplets $(\text{anchor}, \text{positive}, \text{negative})$ are formed at patient level: anchor and positive from the same subject/session (PET$^{+}$) to enforce intra-session consistency, while negatives are cross-subject with CL gap control $|\text{CL}_a - \text{CL}_n| \geq \Delta_{\min} = 5.0$, ensuring negatives are biochemically distinct while remaining challenging to discriminate. If no candidate meets training-set criteria, we apply multi-stage fallback: (i)~progressively relax threshold to $\{2.5, 1.0\}$ and select the patient with maximum CL difference; (ii)~if unsuccessful, sample uniformly from different subjects, preventing identity collisions (different \texttt{subject\_id}) and handling missing values. Negative PET slices are preprocessed identically to positives (resize $224{\times}224$, CLIP normalization) and embedded by the same teacher. Anchor embeddings use \emph{cross-attention} (PET$\rightarrow$MRI), $\mathbf{e}^{\text{anchor}} = \mathrm{AttnPool}(\mathbf{E}_{\text{PET}}^{+},\mathbf{E}_{\text{MRI}})$, to inject MRI context; positives and negatives use \emph{self-attention} on PET$^{+}$ and PET$^{-}$ respectively. This yields clinically-structured hard negatives, improving separability and encouraging fine-grained amyloid discrimination.

\begin{figure}[h!t]
  \centering
  \includegraphics[width=1.00\columnwidth]{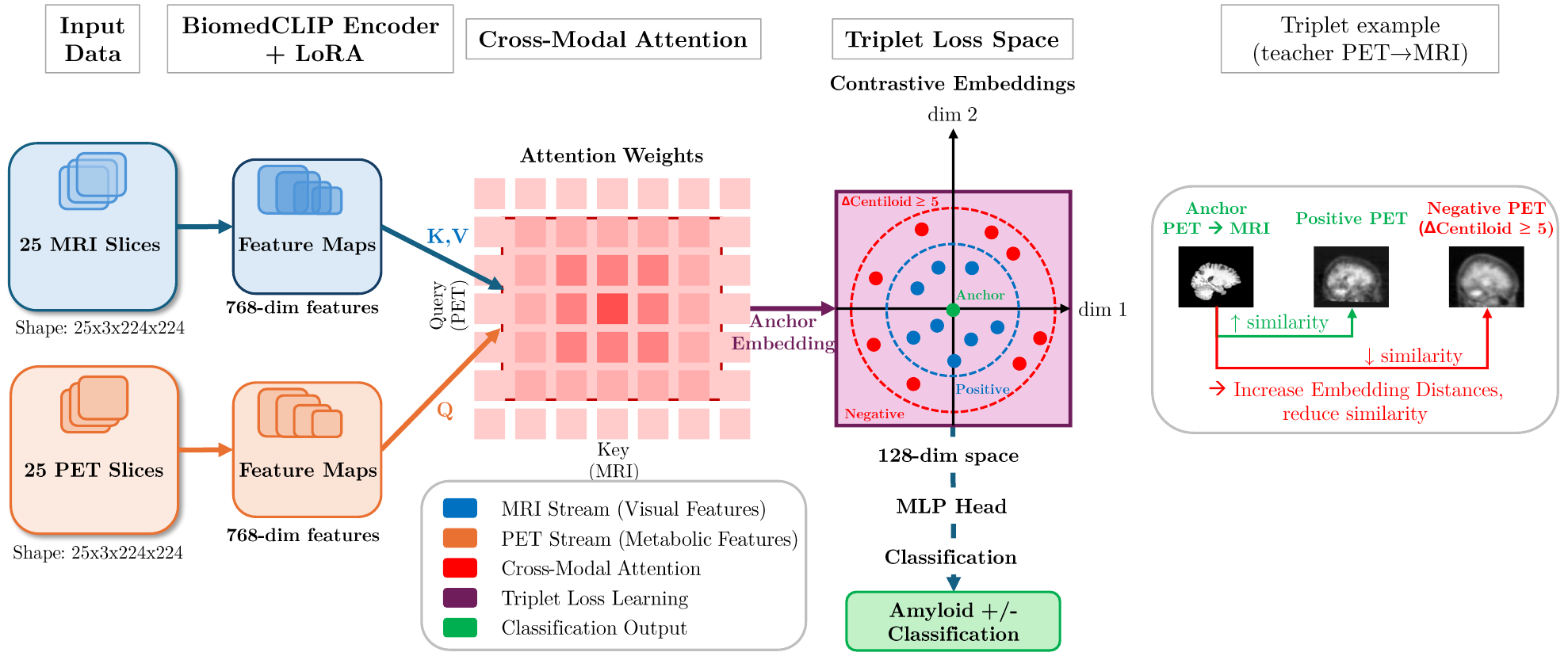}
  \caption{\textbf{Phase 2 --- Teacher refinement} with online CL-aware triplet mining.}
  \label{fig:phase2-triplet}
\end{figure}

\paragraph{Loss Function}
In Phase~2, the training objective combines triplet loss with binary cross-entropy and regularization \textit{(see Fig.~\ref{fig:phase2-triplet})}:
\begin{equation}
\mathcal{L}_{\text{Phase2}} = \mathcal{L}_{\text{triplet}} + \lambda_{\text{cls}} \cdot \text{BCE}(z_T, y) + \mathcal{L}_{\text{reg}},
\end{equation}
where $\mathcal{L}_{\text{triplet}}$ is the triplet loss in Eq.~\eqref{eq:triplet}, $z_T$ is the teacher's classification logit, $y \in \{0,1\}$ is the binary amyloid label, and we use binary cross-entropy with logits (BCE) with $\lambda_{\text{cls}} = 1.0$, and $\mathcal{L}_{\text{reg}}$ includes $\ell_2$ penalty on embedding norms and anti-collapse terms. Mini-batches are balanced via \texttt{WeightedRandomSampler}; we set \texttt{pos\_weight}=1.0 to avoid redundant reweighting. This objective enables the teacher to learn discriminative embeddings via metric learning and accurate classification.

\paragraph{Triplet Loss}

We use triplet loss with margin $m = 1.0$:
\begin{equation}\label{eq:triplet}
\mathcal{L}_{\text{triplet}}
= \tfrac{1}{N}\sum_{i=1}^N
\max\{0,\;\|a_i-p_i\|_2-\|a_i-n_i\|_2+m\},
\end{equation}
where anchor $a = \mathrm{CrossAttn}(\mathbf{E}_{\text{PET}}^+,\mathbf{E}_{\text{MRI}})$, 
positive $p = \mathrm{SelfAttn}(\mathbf{E}_{\text{PET}}^+)$, and 
negative $n = \mathrm{SelfAttn}(\mathbf{E}_{\text{PET}}^-)$ are patient-level 
embeddings after attention pooling. Regularization $\mathcal{L}_{\text{reg}}$ includes $\ell_2$ norm penalty ($\lambda = 0.01$) and adaptive anti-collapse terms. We train for 15 epochs using AdamW with component-specific learning rates: $5 \times 10^{-6}$ (encoder/attention), $2 \times 10^{-5}$ (classifier), and CosineAnnealingWarmRestarts scheduler (full details in supplementary).

\subsection{Phase 3: Knowledge Distillation}
\label{sec:phase3}

\paragraph{MarginFocal Loss}
We use margin-augmented focal loss with margin-shifted logits $\tilde z_i = z_i + m(1-2\hat y_i)$, where $\hat y_i = \mathbf{1}[y'_i > 1/2]$ and $y'_i$ are the (optionally smoothed) targets:

\begin{equation}\label{eq:mf}
\begin{aligned}
\mathcal{L}_{\mathrm{MF}} 
&= \frac{1}{N}\sum_{i=1}^N (1-p_{t_i})^{\gamma}\,\mathrm{BCE}_w(\tilde z_i,y'_i) \\
&\quad + \lambda_{\mathrm{gap}}\,[\,m-(\bar z_{+}-\bar z_{-})\,]_+,
\end{aligned}
\end{equation}
where $\gamma{=}2.0$, $p_i = \sigma(\tilde z_i)$, $p_{t_i} = y'_i p_i + (1{-}y'_i)(1 - p_i)$, and $\bar z_{\pm}$ are batch means of raw logits over positives/negatives. We anneal $m \in [0.3, 1.2]$ over 20 epochs and schedule $\lambda_{\mathrm{gap}} = 0.1$ (epochs 1--10) then $0.3$; gap deficit is internally scaled by $0.1$ (see supplementary).

\paragraph{Feature Distillation}
We align the student's MRI-only embeddings $\hat{\mathbf{e}}_S$ with the teacher's cross-modal embeddings $\hat{\mathbf{e}}_T$ via $\ell_2$-normalized mean squared error:

\begin{equation}
\label{eq:feat_distill}
\mathcal{L}_{\text{feat}} = \left\| \hat{\mathbf{e}}_S - \hat{\mathbf{e}}_T \right\|_2^2,
\quad \hat{\mathbf{e}} = \frac{\mathbf{e}}{\|\mathbf{e}\|_2}.
\end{equation}
\noindent Normalization focuses on directional alignment rather than magnitude; teacher embeddings are detached to ensure unidirectional knowledge transfer.

\begin{figure}[h!t]
  \centering
  \includegraphics[width=1.00\columnwidth]{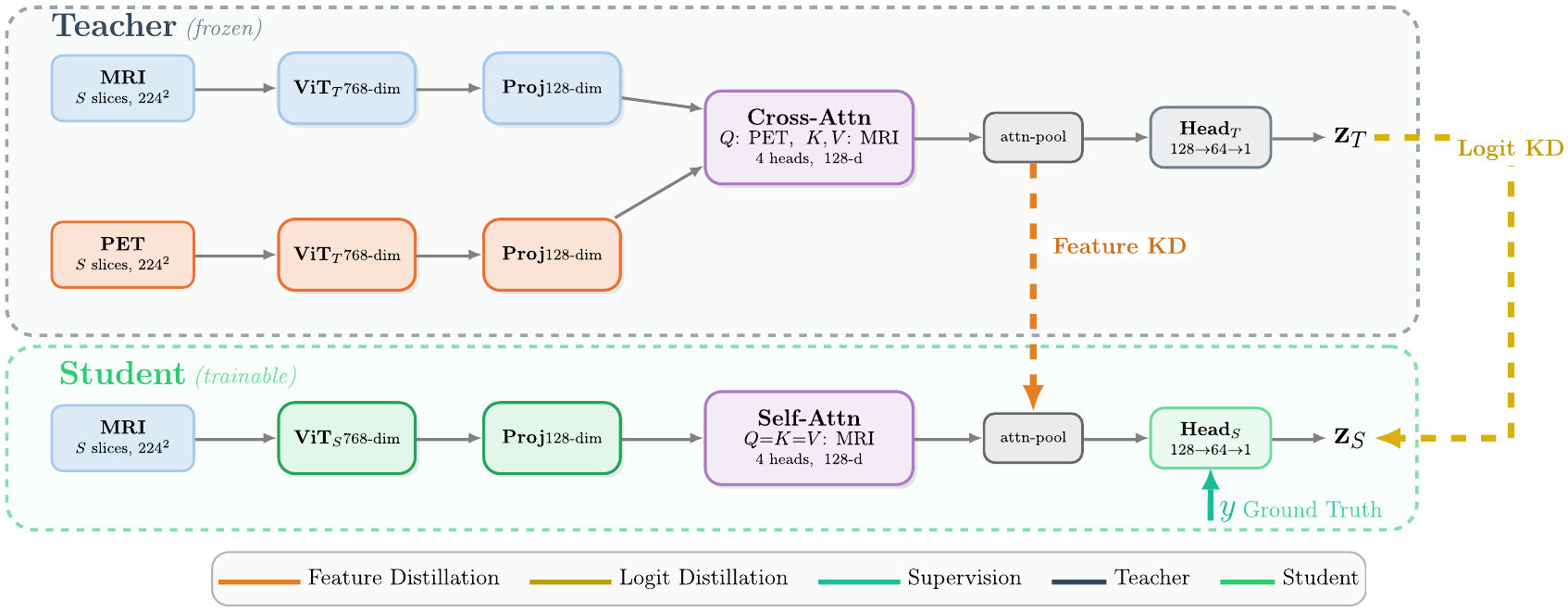}
\caption{\textbf{Phase 3 --- Knowledge distillation}. MRI-only student is trained 
via $\ell_2$ feature matching and temperature-scaled logit distillation; 
PET/CL are used only for teacher supervision.}
  \label{fig:phase3-kd}
\end{figure}

\paragraph{Logit Distillation}
To transfer the teacher’s classification calibration and decision boundaries \textit{(see Fig.~\ref{fig:phase3-kd})}, we apply temperature-scaled logit distillation using binary cross-entropy between the student’s scaled logits and the teacher’s soft targets:
\begin{equation}
\label{eq:logit_distill}
\mathcal{L}_{\text{logit}} = T^2 \cdot \text{BCE}\left(
\frac{z_S}{T}, \sigma\left(\frac{z_T}{T}\right)\right),
\end{equation}
where $T$ is the distillation temperature, $\sigma(\cdot)$ is the sigmoid function, and the $T^2$ factor compensates for the gradient magnitude reduction caused by temperature scaling. Higher temperatures ($T > 1$) produce softer probability distributions enabling the student model to learn from the teacher’s uncertainty and relative confidences rather than only from hard binary labels.

\paragraph{Training Configuration}
\label{paragraph:loss_weight_warmup}
Loss weights warm up linearly over 10 epochs: $\lambda_{\text{cls}}$: $0.3 \to 0.4$, $\lambda_{\text{feat}}$: $0.5 \to 0.4$, $\lambda_{\text{logit}}$: $0.2$, then fixed at $(0.4, 0.4, 0.2)$. Temperature anneals $T = 2.5 \to 1.0$ over 20 epochs. Batch sizes: 6 (Phases 1--2), 10 (Phase 3); gradient accumulation to 30. In Phase 3 we train for 100 epochs using AdamW with component-specific learning rates: LoRA adapters at $2 \times 10^{-4}$ (no wd), projection at $1 \times 10^{-4}$ (wd $10^{-4}$), attention/classifier at $1 \times 10^{-4}$ (wd $10^{-3}$). Scheduler: ReduceLROnPlateau (monitoring val. F1; see supplementary).
 
\subsection{Distillation Loss Function}
\label{sec:distillation-loss}
Our training combines three losses (Eqs.~\eqref{eq:mf}, \eqref{eq:feat_distill}, \eqref{eq:logit_distill}):
\begin{equation}
\label{eq:distillation-loss}
\begin{split}
\mathcal{L}_{\text{total}} = &\lambda_{\text{cls}} \mathcal{L}_{\mathrm{MF}} \\
&+ \lambda_{\text{feat}} \|\hat{\mathbf{e}}_S - \hat{\mathbf{e}}_T\|_2^2 \\
&+ \lambda_{\text{logit}}\, T^2\, \text{BCE}\!\left(\frac{z_S}{T}, \sigma\!\left(\frac{z_T}{T}\right)\right),
\end{split}
\end{equation}
where $\hat{\mathbf{e}} = \text{normalize}(\mathbf{e})$ denotes the $\ell_2$-normalized embeddings after attention pooling, $z$ represents the logit output, and $T$ is the distillation temperature (the $T^2$ factor preserves gradient scale under temperature in binary distillation). Loss weights are annealed as described in Sec.~\ref{paragraph:loss_weight_warmup}~\cite{hintonDistillingKnowledgeNeural2015}.
\section{Experiments}
\label{sec:experiments}
We evaluate our framework on OASIS-3 and ADNI across four widely 
available clinical MRI contrasts (T1w, T2w, FLAIR, T2*) capturing 
complementary tissue properties (anatomy, edema, white-matter lesions, 
microbleeds)~\cite{pelkmansGrayMatterT1w2019,zivanovicRoleMagneticResonance2023}. We first compare with prior MRI-based amyloid prediction methods and present single-modality and multi-contrast results (Sec.~\ref{sec:main_results}, Table~\ref{tab:oasis_adni_joint}), then CDKD (Sec.~\ref{sec:transfer}), and ablations (Sec.~\ref{sec:ablation}). Saliency maps confirm spatially plausible focus patterns in predictions (Sec.~\ref{sec:interpretability}). Checkpoints are selected by validation F1 (\emph{Sel.=F1}) or teacher--student embedding similarity (\emph{Sel.=Sim}). Performance is reported at fixed ($\theta = 0.5$) and validation-optimized ($\theta^*$, maximizing F1) thresholds using standard metrics: F1, accuracy (Acc), precision (Prec), recall (Rec), AUC, and negative predictive value (NPV).

\begin{table*}[!hb]
\centering
\caption{\textbf{Test results on OASIS-3 and ADNI.} \emph{Sel.:} checkpoint by val. F1 or embedding similarity. Metrics: $\theta = 0.5$ vs. optimized $\theta^*$.}
  \label{tab:oasis_adni_joint}
  \setlength{\tabcolsep}{2.6pt}
  \renewcommand{\arraystretch}{1.05}
  {\scriptsize
  \resizebox{\linewidth}{!}{%
  \begin{tabular}{@{}l lcccccccccc lcccccccccc@{}}
    \toprule
    & \multicolumn{11}{c}{\textbf{OASIS-3}} & \multicolumn{11}{c}{\textbf{ADNI}} \\
    \cmidrule(lr){2-12}\cmidrule(l){13-23}
    \multirow{2}{*}{Modality}
      & \multirow{2}{*}{Sel.}
      & \multicolumn{4}{c}{@0.5}
      & \multicolumn{4}{c}{@$\theta^*$}
      & AUC & NPV
      & \multirow{2}{*}{Sel.}
      & \multicolumn{4}{c}{@0.5}
      & \multicolumn{4}{c}{@$\theta^*$}
      & AUC & NPV \\
    & & F1 & Acc & Prec & Rec & F1 & Acc & Prec & Rec & & &
        & F1 & Acc & Prec & Rec & F1 & Acc & Prec & Rec & & \\
    \midrule
    \multicolumn{23}{@{}l}{\textbf{T1w \& T2w available}} \\
    T1w                  & F1  & 0.53 & 0.71 & 0.57 & 0.49 & 0.59 & \textbf{0.71} & 0.56 & 0.64 & 0.73 & 0.81
                         & F1  & 0.58 & 0.56 & 0.48 & 0.74 & \textbf{0.61} & 0.50 & 0.46 & \textbf{0.92} & \textbf{0.66} & 0.77 \\
    T1w                  & Sim & 0.46 & 0.69 & 0.54 & 0.40 & 0.53 & 0.49 & 0.38 & \textbf{0.87} & 0.70 & 0.83
                         & Sim & 0.57 & 0.54 & 0.47 & 0.71 & 0.59 & 0.51 & 0.46 & 0.82 & 0.64 & 0.68 \\
    T2w                  & F1  & 0.54 & 0.54 & 0.40 & 0.80 & 0.55 & 0.63 & 0.46 & 0.69 & 0.70 & 0.80
                         & F1  & 0.51 & 0.63 & 0.59 & 0.45 & 0.57 & 0.51 & 0.45 & 0.76 & 0.62 & 0.65 \\
    T2w                  & Sim & 0.52 & 0.69 & 0.54 & 0.51 & 0.54 & 0.52 & 0.40 & 0.84 & 0.71 & 0.82
                         & Sim & 0.61 & 0.50 & 0.46 & 0.92 & \textbf{0.61} & 0.46 & 0.44 & 1.00 & 0.63 & 1.00 \\
    T1w+T2w$\rightarrow$T1w & F1  & 0.54 & 0.54 & 0.40 & 0.84 & \textbf{0.61} & \textbf{0.71} & 0.54 & 0.71 & \textbf{0.74} & 0.83
                         & F1  & 0.45 & 0.64 & 0.65 & 0.34 & \textbf{0.61} & 0.56 & 0.48 & 0.82 & \textbf{0.66} & 0.73 \\
    T1w+T2w$\rightarrow$T1w & Sim & 0.24 & 0.70 & 0.73 & 0.15 & 0.54 & 0.60 & 0.44 & 0.73 & 0.69 & 0.80
                         & Sim & 0.14 & 0.59 & 0.60 & 0.08 & 0.58 & \textbf{0.59} & 0.51 & 0.68 & 0.65 & 0.69 \\
    T1w+T2w$\rightarrow$T2w & F1  & 0.53 & 0.48 & 0.38 & 0.88 & 0.54 & 0.58 & 0.42 & 0.75 & 0.70 & \textbf{0.84}
                         & F1  & 0.48 & 0.67 & 0.70 & 0.37 & 0.55 & 0.49 & 0.44 & 0.74 & 0.61 & 0.62 \\
    T1w+T2w$\rightarrow$T2w & Sim & 0.47 & 0.68 & 0.52 & 0.42 & 0.54 & 0.57 & 0.42 & 0.78 & 0.69 & 0.81
                         & Sim & 0.51 & 0.62 & 0.56 & 0.47 & 0.53 & 0.50 & 0.44 & 0.66 & 0.60 & 0.61 \\
    \midrule
    \multicolumn{23}{@{}l}{\textbf{FLAIR \& T2* available}} \\
    FLAIR                & F1  & 0.55 & 0.51 & 0.42 & 0.80 & 0.48 & 0.58 & 0.46 & 0.50 & 0.63 & 0.67
                         & F1  & 0.60 & 0.60 & 0.66 & 0.55 & 0.66 & 0.60 & 0.61 & 0.71 & 0.64 & 0.57 \\
    FLAIR                & Sim & 0.51 & 0.65 & 0.55 & 0.47 & 0.53 & \textbf{0.66} & 0.57 & 0.50 & 0.63 & 0.71
                         & Sim & 0.60 & 0.57 & 0.61 & 0.58 & 0.66 & 0.57 & 0.58 & 0.77 & 0.59 & 0.55 \\
    T2*                  & F1  & 0.50 & 0.69 & 0.64 & 0.41 & 0.51 & 0.52 & 0.42 & 0.65 & 0.61 & 0.67
                         & F1  & 0.43 & 0.55 & 0.70 & 0.31 & 0.68 & 0.57 & 0.58 & 0.81 & 0.61 & 0.56 \\
    T2*                  & Sim & 0.50 & 0.45 & 0.38 & 0.71 & 0.52 & 0.53 & 0.43 & 0.68 & 0.58 & 0.69
                         & Sim & 0.51 & 0.57 & 0.68 & 0.40 & \textbf{0.74} & 0.63 & 0.60 & \textbf{0.97} & 0.65 & \textbf{0.86} \\
    FLAIR+T2*$\rightarrow$FLAIR & F1  & 0.49 & 0.70 & 0.68 & 0.38 & 0.53 & 0.52 & 0.42 & 0.71 & 0.63 & 0.69
                         & F1  & 0.68 & 0.58 & 0.58 & 0.81 & 0.67 & 0.56 & 0.57 & 0.83 & 0.61 & 0.53 \\
    FLAIR+T2*$\rightarrow$FLAIR & Sim & 0.46 & 0.66 & 0.59 & 0.38 & 0.52 & 0.63 & 0.51 & 0.53 & \textbf{0.64} & 0.70
                         & Sim & 0.69 & 0.61 & 0.61 & 0.79 & 0.73 & \textbf{0.64} & 0.62 & 0.89 & 0.67 & 0.71 \\
    FLAIR+T2*$\rightarrow$T2*   & F1  & 0.47 & 0.67 & 0.62 & 0.38 & 0.56 & 0.54 & 0.44 & 0.77 & 0.63 & 0.82
                         & F1  & 0.71 & 0.56 & 0.56 & 0.96 & 0.71 & 0.56 & 0.56 & 0.95 & 0.59 & 0.64 \\
    FLAIR+T2*$\rightarrow$T2*   & Sim & 0.54 & 0.48 & 0.41 & 0.79 & \textbf{0.57} & 0.43 & 0.40 & 1.00 & 0.63 & 1.00
                         & Sim & 0.70 & 0.63 & 0.62 & 0.80 & 0.71 & 0.56 & 0.56 & 0.96 & \textbf{0.68} & 0.64 \\
\bottomrule
\end{tabular}
}
}
\end{table*}

\subsection{Performance on OASIS-3 and ADNI}
\label{sec:main_results}

\begin{figure}[!hb]
  \centering
  \makebox[\columnwidth][c]{%
    \begin{subfigure}[b]{0.52\columnwidth}
      \includegraphics[width=1.00\columnwidth]{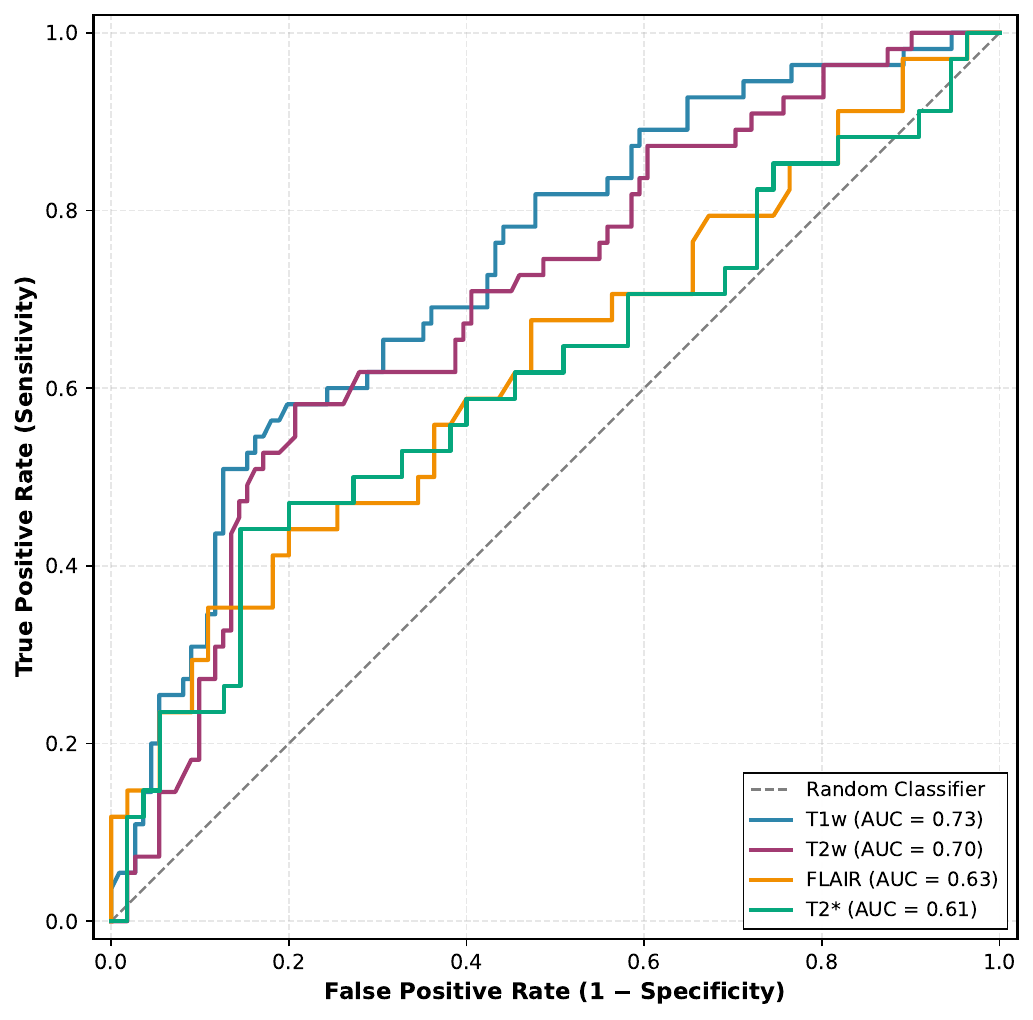}
      \caption{OASIS-3}
    \end{subfigure}%
    \hspace{0.01\columnwidth}%
    \begin{subfigure}[b]{0.52\columnwidth}
      \includegraphics[width=1.00\columnwidth]{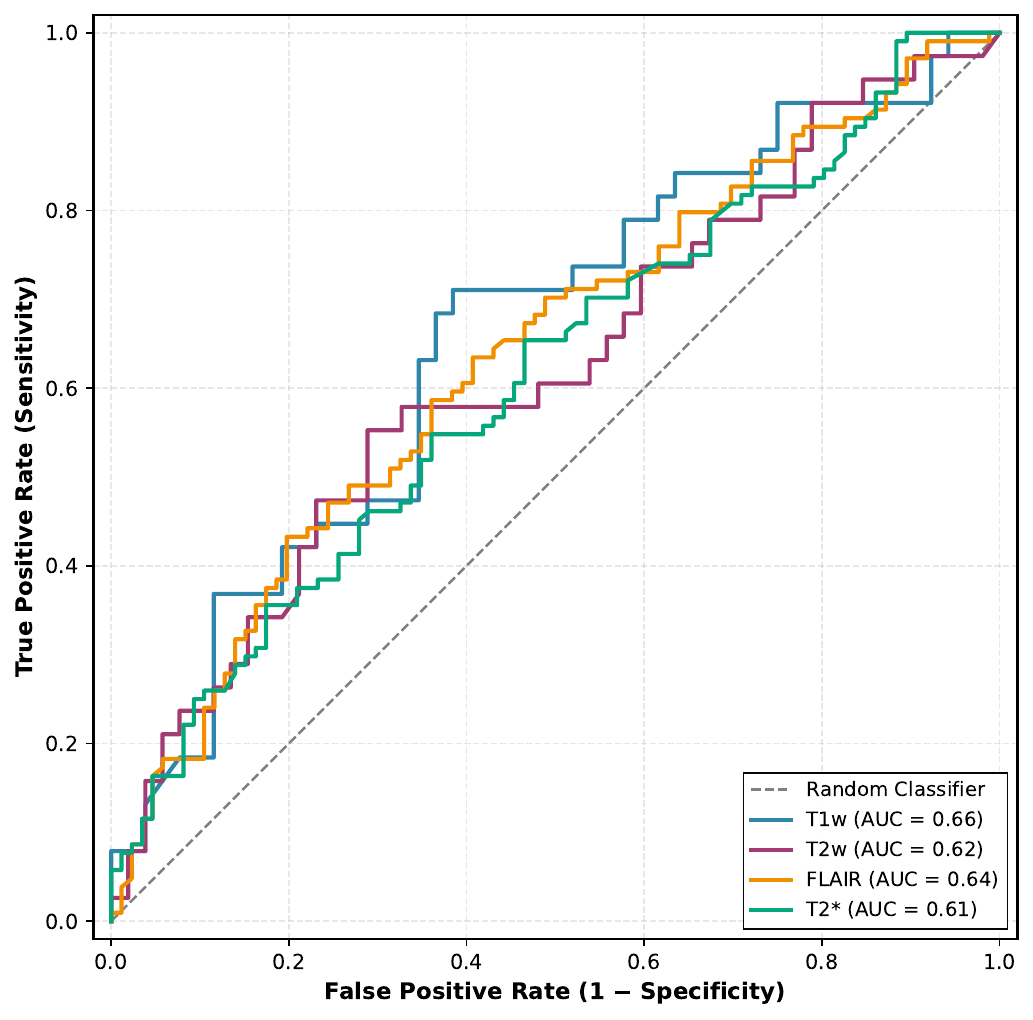}
      \caption{ADNI}
    \end{subfigure}%
  }
  \caption{\textbf{ROC curves per sequence.} (a) OASIS-3; (b) ADNI.}
  \label{fig:roc_comparison}
\end{figure}

Table~\ref{tab:oasis_adni_joint} presents results across both cohorts. T1w achieves the strongest single-sequence performance on OASIS-3 (AUC 0.73 [0.66--0.82]) with balanced metrics at $\theta^*$ (F1 0.59, Acc 0.71, NPV 0.81). T2w attains slightly lower AUC (0.70) but comparable F1 (0.55). 
ADNI results confirm consistency (T1w: AUC 0.66; T2w: 0.62). PET-guided fusion improves MRI-only inference: T1w+T2w$\rightarrow$T1w reaches AUC 0.74 on OASIS-3 while preserving ADNI performance (F1 0.61, Rec 0.82 at $\theta^*$). \emph{Sel.=F1} generally outperforms \emph{Sel.=Sim} for single-contrast students, but \emph{Sel.=Sim} benefits multi-contrast models, confirming PET-aligned embeddings stabilize cross-contrast integration. Fig.~\ref{fig:roc_comparison} reveals consistent ranking: T1w$>$T2w/FLAIR$>$T2*.

\noindent FLAIR and T2* show moderate discrimination, likely due to smaller sample sizes and weaker structural correlates, yet remain useful at $\theta^*$ for high-recall screening.
 
\paragraph{Comparison with Prior Work}
\label{sec:comparison}

Table~\ref{tab:comparison_lew} compares with prior work. Relative to Kim et
al.~\cite{kimDeepLearningBasedPrediction2025}, a large MRI-only baseline, our T1w improves AUC from 0.61 to 0.73 (+19.7\%) and multi-contrast fusion (T1w+T2w: AUC=0.74 vs.\ 0.67 for T1w+FLAIR, +10.4\%, without non-imaging covariates as in Wang et al.~\cite{wangDeeplearningBasedMultimodal2025}). Key advances include: (1)~multi-contrast fusion, (2)~CDKD (ADNI$\leftrightarrow$OASIS-3), and (3)~interpretability via saliency and anatomical bias control (Sec.~\ref{sec:interpretability}). Compared to Chiumento et al.~\cite{chiumentoDetectingBetaAmyloidCrossModal2025} (T1w: F1=0.48), PET-guided Centiloid-aware triplet distillation achieves F1=0.59 (+22.9\%) extending to CL-standardized multi-contrast evaluation.

\begin{table}[!hb]
\centering
\caption{\textbf{Comparison with MRI-based amyloid prediction}. MRI-only inference prioritized; covariate-based method noted.}
\label{tab:comparison_lew}
\setlength{\tabcolsep}{3.5pt}
\resizebox{\columnwidth}{!}{%
\begin{tabular}{@{}lcccccc@{}}
\toprule
\textbf{Method} & \textbf{MRI Input} & \textbf{Ground Truth} & \textbf{Clin.} & \textbf{Training} & \textbf{N} & \textbf{AUC [95\% CI]} \\
\midrule
Kim et al. (2025)~\cite{kimDeepLearningBasedPrediction2025} (T1w)
  & \textbf{T1w} & \textbf{PET (CL)} & \textbf{No} & Direct & 779 & 0.61 [0.59--0.63] \\
Kim et al. (2025)~\cite{kimDeepLearningBasedPrediction2025} (T1w+FLAIR)
  & T1w+FLAIR & \textbf{PET (CL)} & \textbf{No} & Direct & 779 & 0.67 [0.65--0.70] \\
Wang et al. (2025)~\cite{wangDeeplearningBasedMultimodal2025}
  & T1w & PET (SUVR${>}$1.11) & Yes$^\dag$ & Transfer & 241 & 0.74 \\
Dolci et al. (2025)~\cite{dolciMultimodalMRIAccurately2025} 
  & 3-modal & CSF (A$\beta$42)$^\ddagger$ & \textbf{No} & Direct & -- & -- \\
Chiumento et al. (2025)~\cite{chiumentoDetectingBetaAmyloidCrossModal2025}
  & \textbf{T1w} & \textbf{PET (CL)} & \textbf{No} & Direct & -- & -- \\
\midrule
\textbf{Ours T1w} 
  & \textbf{T1w} & \textbf{PET (CL)} & \textbf{No} & \textbf{KD+Trip} & 166 & \textbf{0.73 [0.66--0.82]} \\
\textbf{Ours T2w} 
  & \textbf{T2w} & \textbf{PET (CL)} & \textbf{No} & \textbf{KD+Trip} & 166 & \textbf{0.70 [0.61--0.78]} \\
\textbf{Ours T1w+T2w} 
  & \textbf{T1w+T2w} & \textbf{PET (CL)} & \textbf{No} & \textbf{KD+Trip} & 166 & \textbf{0.74 [0.67--0.82]} \\
\bottomrule
\end{tabular}%
}
\begin{tablenotes}
\footnotesize
\item $^\dag$ Requires non-imaging covariates (age, sex, APOE genotype).
\item -- not reported. Dolci: Acc=0.76; Chiumento: F1=0.48 ($\theta^*$).
\item $^\ddagger$ CSF-based ground truth; not directly comparable to PET-CL.
\item N = test set; CL = Centiloid; CI = conf. interval (bootstrap $1{,}000\times$).
\end{tablenotes}
\end{table}

\subsection{Interpretability Analysis}
\label{sec:interpretability}
\begin{figure}[!t]
\centering
\includegraphics[width=1.00\columnwidth]{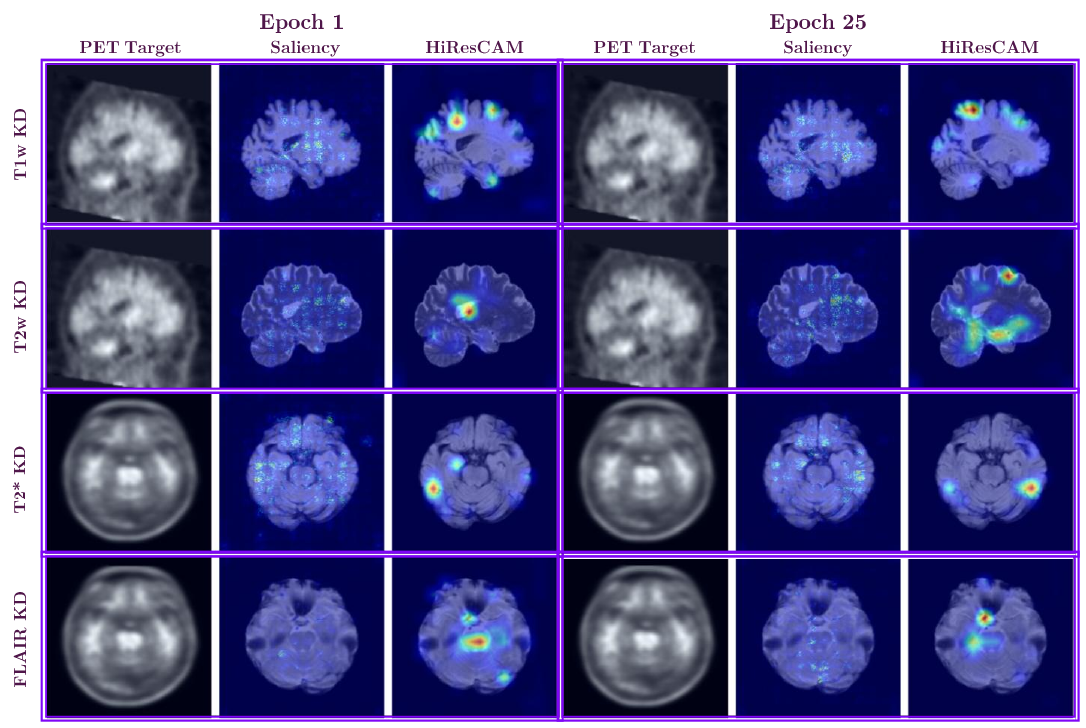}
\caption{\textbf{Saliency/HiResCAM.} OASIS-3, epochs \(1\to 25\).}
\label{fig:saliency_progression}
\end{figure}

Fig.~\ref{fig:saliency_progression} shows saliency/HiResCAM maps across training (epochs 1$\to$25) on OASIS-3 for all MRI contrasts~\cite{mahmudExplainableAIParadigm2024,draelosUseHiResCAMInstead2021,selvarajuGradCAMVisualExplanations2017}. At epoch 1, attention is diffuse; by epoch 25, the model focuses on regions consistent across contrasts, with spatial alignment between MRI and PET evident in HiResCAM.

\subsection{Cross-Dataset Transfer}
\label{sec:transfer}
Table~\ref{tab:transfer_teacher} evaluates cross-dataset knowledge distillation (source teacher distills target MRI-only student)~\cite{martenssonReliabilityDeepLearning2020a,bentoDeepLearningLarge2022}. \textbf{ADNI$\rightarrow$OASIS-3:} ADNI teachers transfer effectively (AUC 0.72--0.73). T2w transfer outperforms native training (T2w: 0.72 vs. 0.70; T1w+T2w$\rightarrow$T2w: 0.73 vs. 0.70, Sel.=F1). T1w is stable ($\Delta\text{AUC} < 0.01$), while T1w+T2w$\rightarrow$T1w reaches F1 0.61 and NPV 0.83 at $\theta^*$. \textbf{OASIS-3$\rightarrow$ADNI:} T1w transfers best (AUC 0.66, F1 0.60 at $\theta^*$). Fusion preserves or improves NPV under $\theta^*$; \emph{Sel.=F1} balances metrics while \emph{Sel.=Sim} attains higher recall. This reflects target-dataset adaptation in realistic clinical settings with limited paired PET--MRI data.
\begin{table}[!hbt]
  \centering
  \caption{\textbf{Cross-dataset knowledge distillation (CDKD).} Source teacher distills target MRI-only student. Metrics at $\theta = 0.5$, $\theta^*$.}
  \label{tab:transfer_teacher}
  \setlength{\tabcolsep}{2.6pt} 
  \renewcommand{\arraystretch}{1.05}
  {\scriptsize
  \begin{tabular}{lcccccccccc}
    \toprule
    & \multicolumn{4}{c}{@0.5} & \multicolumn{4}{c}{@$\theta^*$} & \multicolumn{2}{c}{Overall} \\
    \cmidrule(lr){2-5}\cmidrule(lr){6-9}\cmidrule(l){10-11}
    Modality & F1 & Acc & Prec & Rec & F1 & Acc & Prec & Rec & AUC & NPV \\
    \midrule
    \multicolumn{11}{@{}l}{\textbf{ADNI (Teacher) KD$\rightarrow$ OASIS-3 (Student)} --- \emph{Sel.=F1}} \\
    T1w                  & 0.27 & 0.68 & 0.53 & 0.18 & 0.58 & \textbf{0.70} & 0.54 & 0.64 & \textbf{0.73} & 0.80 \\
    T2w                  & 0.55 & 0.70 & 0.54 & 0.56 & 0.54 & 0.69 & 0.54 & 0.55 & 0.72 & 0.77 \\
    T1w+T2w$\rightarrow$T1w & 0.55 & 0.58 & 0.42 & 0.76 & \textbf{0.61} & 0.70 & 0.53 & 0.71 & 0.72 & \textbf{0.83} \\
    T1w+T2w$\rightarrow$T2w & 0.39 & 0.72 & 0.68 & 0.27 & 0.57 & 0.61 & 0.45 & \textbf{0.76} & \textbf{0.73} & 0.82 \\
    \midrule
    \multicolumn{11}{@{}l}{\textbf{OASIS-3 (Teacher) $\rightarrow$ ADNI (Student)} --- \emph{Sel.=F1}} \\
    T1w                  & 0.54 & 0.63 & 0.58 & 0.50 & \textbf{0.60} & \textbf{0.61} & 0.53 & 0.68 & \textbf{0.66} & \textbf{0.71} \\
    T2w                  & 0.49 & 0.60 & 0.53 & 0.45 & 0.57 & 0.52 & 0.46 & \textbf{0.74}& 0.61 & 0.66 \\
    T1w+T2w$\rightarrow$T1w & 0.46 & 0.60 & 0.54 & 0.40 & 0.56 & 0.53 & 0.47 & 0.71 & 0.62 & 0.66 \\
    T1w+T2w$\rightarrow$T2w & 0.10 & 0.60 & 1.00 & 0.05 & 0.55 & 0.51 & 0.45 & 0.71 & 0.62 & 0.63 \\
    \bottomrule
  \end{tabular}
  }
\end{table}

\subsection{Ablation Study}
\label{sec:ablation}
Table~\ref{tab:ablation} shows T1w/T2w ablations. \textbf{Phase 1 (Pre-training):} Omitting pre-training lowers OASIS-3 T1w F1 from 0.59 to 0.53 at $\theta^*$, with Prec./Rec. imbalance (0.46/0.75 at $\theta = 0.5$), indicating miscalibration. Larger drop on OASIS-3 (0.59 $\to$ 0.53) than ADNI (0.61 $\to$ 0.59), likely due to class imbalance. \textbf{Phase 2 (Triplet):} Removing triplet mining drops OASIS-3 T1w F1 to 0.51 ($\Delta$F1 $=-0.08$), showing CL-aware hard-negative separation. \textbf{Phase 3A:} Removing embedding alignment yields the largest drop (OASIS-3 T1w: 0.51; T2w: 0.52 at $\theta^*$), lowering T2w NPV to 0.78 (from 0.80--0.82), confirming PET--MRI alignment is critical. \textbf{Phase 3B:} Removing logit distillation yields intermediate performance (T1w: 0.55; T2w: 0.52), worsening calibration. \textbf{Ranking:} 3A (feature) yields largest F1 gains; 1 (pre-training) helps imbalanced data; 2 (hard negatives) improves discrimination; 3B (logit) refines calibration.

\begin{table}[!hbt]
  \centering
\caption{\textbf{Ablation Study.} T1w/T2w phase impact ($\theta = 0.5$, $\theta^*$).}
  \label{tab:ablation}
  \setlength{\tabcolsep}{2.6pt}
  \renewcommand{\arraystretch}{1.05}
  {\scriptsize
  \begin{tabular}{@{}llcccccccccc@{}}
    \toprule
    \multicolumn{2}{@{}l}{} & \multicolumn{4}{c}{@0.5} & \multicolumn{4}{c}{@$\theta^*$} & \multicolumn{2}{c}{Overall} \\
    \cmidrule(lr){3-6}\cmidrule(lr){7-10}\cmidrule(l){11-12}
    Dataset & Modality & F1 & Acc & Prec & Rec & F1 & Acc & Prec & Rec & AUC & NPV \\
    \midrule
    \multicolumn{12}{@{}l}{\textbf{No Pre-training (Phase 1)}} \\
    OASIS-3 & T1w & 0.57 & 0.62 & 0.46 & 0.75 & 0.53 & 0.71 & 0.56 & 0.51 & 0.74 & 0.77 \\
    OASIS-3 & T2w & 0.16 & 0.68 & 0.56 & 0.09 & 0.55 & 0.62 & 0.45 & 0.71 & 0.67 & 0.80 \\
    ADNI    & T1w & 0.57 & 0.57 & 0.49 & 0.68 & 0.59 & 0.57 & 0.49 & 0.73 & 0.63 & 0.70 \\
    ADNI    & T2w & 0.51 & 0.57 & 0.49 & 0.53 & 0.55 & 0.50 & 0.44 & 0.74 & 0.61 & 0.63 \\
    \midrule
    \multicolumn{12}{@{}l}{\textbf{No Triplet Loss (Phase 2)}} \\
    OASIS-3 & T1w & 0.39 & 0.72 & 0.68 & 0.27 & 0.51 & 0.70 & 0.55 & 0.47 & 0.71 & 0.76 \\
    OASIS-3 & T2w & 0.55 & 0.61 & 0.44 & 0.71 & 0.56 & 0.60 & 0.44 & 0.78 & 0.70 & 0.82 \\
    ADNI    & T1w & 0.62 & 0.67 & 0.60 & 0.63 & 0.60 & 0.59 & 0.51 & 0.74 & 0.68 & 0.71 \\
    ADNI    & T2w & 0.47 & 0.64 & 0.64 & 0.37 & 0.57 & 0.52 & 0.46 & 0.76 & 0.60 & 0.67 \\
    \midrule
    \multicolumn{12}{@{}l}{\textbf{No Feature Distillation (Phase 3A)}} \\
    OASIS-3 & T1w & 0.43 & 0.70 & 0.58 & 0.35 & 0.51 & 0.68 & 0.52 & 0.51 & 0.71 & 0.76 \\
    OASIS-3 & T2w & 0.53 & 0.66 & 0.49 & 0.58 & 0.52 & 0.56 & 0.41 & 0.73 & 0.71 & 0.78 \\
    ADNI    & T1w & 0.60 & 0.56 & 0.48 & 0.79 & 0.61 & 0.50 & 0.46 & 0.92 & 0.65 & 0.77 \\
    ADNI    & T2w & 0.50 & 0.58 & 0.50 & 0.50 & 0.56 & 0.50 & 0.45 & 0.76 & 0.61 & 0.64 \\
    \midrule
    \multicolumn{12}{@{}l}{\textbf{No Logit Distillation (Phase 3B)}} \\
    OASIS-3 & T1w & 0.61 & 0.66 & 0.49 & 0.78 & 0.55 & 0.71 & 0.57 & 0.53 & 0.74 & 0.77 \\
    OASIS-3 & T2w & 0.54 & 0.63 & 0.46 & 0.66 & 0.52 & 0.55 & 0.40 & 0.73 & 0.69 & 0.78 \\
    ADNI    & T1w & 0.63 & 0.62 & 0.54 & 0.76 & 0.60 & 0.53 & 0.47 & 0.84 & 0.67 & 0.72 \\
    ADNI    & T2w & 0.49 & 0.58 & 0.50 & 0.47 & 0.57 & 0.51 & 0.45 & 0.76 & 0.60 & 0.65 \\
    \bottomrule
  \end{tabular}
  }
\end{table}

\section{Conclusion}
We introduce a PET-guided knowledge distillation framework for MRI-only amyloid-$\beta$ prediction, with no clinical or demographic variables at inference. Unlike prior approaches dependent on genetic or cognitive covariates, our method achieves competitive performance using standard MRI alone, enhancing deployability in settings with high data missingness. A BiomedCLIP teacher, trained via Centiloid-aware triplet mining, transfers knowledge to an MRI-only student network. Evaluations on OASIS-3 and ADNI across four MRI contrasts show strong cross-dataset transfer with CDKD. Ablations confirm that both feature distillation and triplet learning are critical, with pre-training providing additional gains under data imbalance. Saliency analyses reveal anatomically plausible regions contributing to prediction. Overall, this framework offers a promising path toward cost-effective, scalable Alzheimer’s biomarker screening using routine MRI. Future work will explore multi-contrast fusion beyond tested pairs, advanced MRI (diffusion, perfusion), and prognostic prediction of amyloid conversion using these sequences.

\section*{Acknowledgments}
This work was funded by Taighde \'Eireann -- Research Ireland through the Research Ireland Centre for Research Training in Machine Learning (18/CRT/6183). This work was also supported by Taighde \'Eireann -- Research Ireland under Grant number SFI/12/RC/2289\_P2, co-funded by the European Regional Development Fund through the Research Ireland Insight Centre for Data Analytics at Dublin City University.

\ADNIAckText

\OASISAckText

{
    \small
    \bibliographystyle{ieeenat_fullname}
    \bibliography{CVPR_2026_Workshop}
}

\appendix
\maketitlesupplementary

\section{List of Abbreviations}
\label{suppl:abbreviations}

Table~\ref{tab:abbreviations} summarizes the main abbreviations used throughout the main paper and supplementary material.

\begin{table}[!htb]
\centering
\caption{\textbf{Abbreviations used throughout the main paper and supplementary material.}}
\label{tab:abbreviations}
\scriptsize
\renewcommand{\arraystretch}{1.1}
\resizebox{\columnwidth}{!}{
\begin{tabular}{@{}llll@{}}
\toprule
\textbf{Abbr.} & \textbf{Full Form} & \textbf{Abbr.} & \textbf{Full Form} \\
\midrule
\multicolumn{4}{@{}l}{\textit{Medical \& Imaging}} \\
A$\beta$ & Amyloid-$\beta$ & OASIS-3 & Open Access Series of Imaging Studies 3 \cite{lamontagneOASIS3LongitudinalNeuroimaging2019d}\\
AD & Alzheimer's disease & PET & Positron Emission Tomography \\
ADNI & Alzheimer's Disease Neuroimaging Initiative \cite{alzheimersdiseaseneuroimaginginitiativeadniAlzheimersDiseaseNeuroimaging2025} & PiB & Pittsburgh Compound B \cite{yaminPittsburghCompoundBPiB2017} \\
CL & Centiloid & PUP & PET Unified Pipeline \cite{suYsu001PUP2025}\\
CSF & Cerebrospinal Fluid & SUVR & Standardized Uptake Value Ratio \\
FLAIR & Fluid-Attenuated Inversion Recovery & T1w & T1-weighted \\
T2w & T2-weighted & T2* & T2*-weighted \\
FWHM & Full Width at Half Maximum & NFTs & Neurofibrillary Tangles \\
MRI & Magnetic Resonance Imaging & APOE4 & Apolipoprotein E $\varepsilon 4$ allele \\
APOE & Apolipoprotein E & AV-45 & Florbetapir ($^{18}$F-AV-45) \\
ATN & Amyloid--Tau--Neurodegeneration framework & SWI & Susceptibility-Weighted Imaging \\
ARIA & Amyloid-Related Imaging Abnormalities & ICBM152 & MNI ICBM152 brain template \cite{fonovUnbiasedNonlinearAverage2009} \\
DTI & Diffusion Tensor Imaging & EHR & Electronic Health Record \\
GRE & Gradient-Recalled Echo & & \\

\midrule
\multicolumn{4}{@{}l}{\textit{Deep Learning \& Architecture}} \\
BCE & Binary Cross-Entropy & MLP & Multi-Layer Perceptron \\
CLS & Classification Token & ReLU & Rectified Linear Unit \\
GELU & Gaussian Error Linear Unit & ViT & Vision Transformer \\
KD & Knowledge Distillation & MHA & Multi-Head Attention \\
LoRA & Low-Rank Adaptation \cite{huLoRALowRankAdaptation2021c}& CDKD & Cross-Dataset Knowledge Distillation \\
CLIP & Contrastive Language--Image Pre-training & BiomedCLIP & Biomedical CLIP vision--language model \\
\midrule
\multicolumn{4}{@{}l}{\textit{Preprocessing \& Tools}} \\
ANTs & Advanced Normalization Tools \cite{tustisonANTsXEcosystemQuantitative2021}& MNI & Montreal Neurological Institute \cite{fonovUnbiasedNonlinearAverage2009}\\
HD-BET & HD Brain Extraction Tool \cite{isenseeAutomatedBrainExtraction2019a} & N4 & N4 Bias Field Correction \cite{tustisonN4ITKImprovedN32010}\\
\midrule
\multicolumn{4}{@{}l}{\textit{Evaluation Metrics}} \\
Acc & Accuracy & F1 & F1 score \\
AUC & Area Under the ROC Curve & Prec & Precision \\
Rec & Recall & NPV & Negative Predictive Value \\
CI & Confidence Interval & ROC & Receiver Operating Characteristic \\
\midrule
\multicolumn{4}{@{}l}{\textit{Training}} \\
AdamW & Adam with Weight Decay \cite{loshchilovDecoupledWeightDecay2019}& lr & Learning Rate \\
FP16 & 16-bit Floating Point & wd & Weight Decay \\
\bottomrule
\end{tabular}
}
\end{table}

\section{Supplementary Material Overview}
\label{suppl:overview}

In this section, we provide technical details and additional experiments that support the main paper. In Sec.~\ref{suppl:implementation}, we present further implementation details, including augmentation strategies, hyperparameters for all three training phases, and architecture-specific formulas referenced in the main text. In Sec.~\ref{suppl:additional_experiments}, we report additional qualitative and quantitative results: radar charts to visualize performance trade-offs across metrics and to compare single-sequence models with multi-sequence distilled models (Fig.~\ref{fig:radar_comparison}); ablation studies evaluating different Centiloid thresholds for negative patient selection during triplet mining (Sec.~\ref{suppl:ablation_centiloid}, Table~\ref{tab:ablation_centiloid}); ROC curve evolution from the first to the last epoch (Sec.~\ref{suppl:roc_analysis}, Fig.~\ref{fig:roc_combined}); and interpretability analyses (Sec.~\ref{suppl:interpretability}) for both single- and multi-sequence models via gradient-based saliency~\cite{simonyan2013deep} and HiResCAM~\cite{draelosUseHiResCAMInstead2021}, which confirm that the model's attention focuses on anatomically plausible regions (Figs.~\ref{fig:suppl_interp_combined}--\ref{fig:suppl_interp_multi_combined}).

\section{Implementation Details}
\label{suppl:implementation}

\subsection{Data Augmentation}
\label{suppl:data_augmentation}

To improve generalization while preserving PET--MRI spatial correspondence and to facilitate reproducibility, we apply the following synchronized augmentations during Phases~1--2 using shared random seeds for each PET--MRI pair:

\begin{itemize}
    \item \textbf{Spatial transformations:} random affine (rotation $\pm 7^\circ$),
    translation $\pm 5\%$, and isotropic scaling in $[0.95, 1.05]$.
    \item \textbf{Intensity modulations:} color jitter (brightness/contrast $\pm 10\%$),
    Gaussian blur (kernel size 3, $p = 0.3$), gamma correction
    $\gamma \in [0.9, 1.1]$ ($p = 0.5$), and Gaussian noise
    $\sigma \in [0.01, 0.03]$ ($p = 0.5$).
    \item \textbf{Random erasing:} $p = 0.25$, erase scale in $[0.05, 0.12]$.
\end{itemize}

\noindent The same augmentations are applied during Phase 3, with PET--MRI synchronization preserved (shared random seeds). At test time, only resizing and normalization are used.

\noindent\textbf{Framework \& hardware.}
We use PyTorch~2.x with CUDA~12.x on NVIDIA GeForce RTX~4090/5090 GPUs (24--32~GB VRAM).

\subsection{Training Configuration}
\label{suppl:training_config}
We use PyTorch with seed $= 42$, mixed-precision training (FP16), and gradient accumulation to increase the effective batch size. Hyperparameters are tuned on the validation set and then kept fixed for all experiments. Early stopping is based on validation F1 (Phases~1 and 3) and on a combined score (triplet separation + F1) in Phase~2.

\paragraph{Class Balancing and Sampling}
\label{suppl:class_balance}
To mitigate label imbalance, we use per-class inverse-frequency weights and a weighted sampler. Let $n_0, n_1$ be the counts of negative/positive samples in the training set; we set per-class weights $w_c = 1/n_c$ and assign each sample the weight of its class. A \texttt{WeightedRandomSampler} (replacement) is used to draw mini-batches with balanced label proportions. We do not apply additional positive reweighting in the BCE (i.e., \texttt{pos\_weight} $=1.0$) to avoid double-counting the imbalance already handled by sampling.

\noindent\textbf{Reproducibility.}
Fixed seed (42) for all stochastic operations (data loading, augmentation, model initialization, dropout); CuDNN in deterministic mode with benchmarking disabled.

\noindent\textbf{Training:} Mixed precision with \texttt{GradScaler} and gradient accumulation; batch sizes 6 (Phases~1--2), 10 (Phase~3).

\noindent\textbf{Early Stopping:}
\begin{itemize}
    \item Phase 1 (Pre-training): validation F1, patience 5
    \item Phase 2 (Contrastive): validation combined score (triplet separation + F1), patience 3
    \item Phase 3 (Distillation): validation F1, patience 25
\end{itemize}

\noindent\textbf{Data loading \& checkpointing:}
4 DataLoader workers with pin memory enabled. Phase~1 (pre-training): best validation F1. Phase~2 (teacher): best \emph{combined score} (triplet separation + classification F1, weight 1:0.5). Phase~3 (student): best teacher--student similarity and best validation F1 (saved separately).

\subsection{Architecture Implementation Details}
\label{suppl:architecture_details}

\subsubsection{LoRA Parameterization}
\label{suppl:lora_details}
Low-Rank Adaptation (LoRA)~\cite{huLoRALowRankAdaptation2021c} reparameterizes each projection with pretrained weights $\mathbf{W}_0\in\mathbb{R}^{d_{\text{out}}\times d_{\text{in}}}$ as:
\begin{equation}
\mathbf{W}' = \mathbf{W}_0 + s\,\Delta\mathbf{W}, \quad \Delta\mathbf{W}=\mathbf{B}\mathbf{A}, \quad s=\alpha/r,
\end{equation}

\noindent where $\mathbf{B}\in\mathbb{R}^{d_{\text{out}}\times r}$ and $\mathbf{A}\in\mathbb{R}^{r\times d_{\text{in}}}$ are the only trainable parameters, with rank $r = 32$ and scaling factor $\alpha = 32$ (so $s = \alpha/r = 1.0$), while $\mathbf{W}_0$ remains frozen. We rely on the standard LoRA initialization from the PEFT library: the up-projection $\mathbf{B}$ is initialized to zero so that $s\,\mathbf{B}\mathbf{A} = \mathbf{0}$ at the beginning of training, and $\mathbf{W}' = \mathbf{W}_0$. LoRA adapters are applied to attention projections (query, key, value, output) in transformer blocks 6--11 (0-indexed).

\subsubsection{Projection Head Architecture}
\label{suppl:projhead_arch}
ViT CLS tokens (768D) are projected to 128D using the dropout rates specified in Table~\ref{tab:dropout_schedules}:
\begin{equation}
\begin{aligned}
\mathbf{h}^{(1)} &= \text{Dropout}_1\bigl(\text{GELU}(\text{LN}_1(\mathbf{W}_1 \mathbf{h}))\bigr),\\
\mathbf{e}       &= \text{Dropout}_2\bigl(\text{LN}_2(\mathbf{W}_2 \mathbf{h}^{(1)})\bigr).
\end{aligned}
\label{eq:projection_head}
\end{equation}

\noindent where $\mathbf{W}_1: 768 \to 256$ and $\mathbf{W}_2: 256 \to 128$.

\subsubsection{Attention Pooling Formula}
\label{suppl:pooling_formula}
Given slice embeddings $\mathbf{A} = [\mathbf{A}_1, \dots, \mathbf{A}_S]\in\mathbb{R}^{S\times128}$ after MHA (4 heads, $d_{\text{head}} = 32$), with $S$ slices, we compute:
\begin{equation}
w_s = \mathbf{W}_{\text{pool}}\mathbf{A}_s + b, \quad \alpha_s = \text{softmax}(w_s/\tau),
\end{equation}
where $\mathbf{W}_{\text{pool}}\in\mathbb{R}^{1\times 128}$, $b \in \mathbb{R}$ is a scalar bias term, and $\tau = 2.0$. The patient representation is $\mathbf{e} = \sum_{s=1}^S \alpha_s\,\mathbf{A}_s$.

\subsubsection{Classification Head Architecture}
\label{suppl:cls_head_details}
Both teacher and student use:
\begin{equation}
z = \mathbf{W}_2(\text{Dropout}(\text{ReLU}(\mathbf{W}_1 \mathbf{e}))),
\end{equation}
with $\mathbf{W}_1: 128 \to 64$, $\mathbf{W}_2: 64 \to 1$. 

\noindent\textbf{Weight initialization:} the attention pooling weight uses Xavier-uniform (gain = 1.0) with bias 0.0; other linear layers use PyTorch defaults. Before student distillation, the classifier output layer ($64{\to} 1$) is re-initialized with Xavier-uniform (gain = 0.5) and bias 0.0.

\begin{table}[!htb]
\centering
\scriptsize
\renewcommand{\arraystretch}{0.95}
\caption{\textbf{Dropout schedules.}
Dropout rates for projection and classification heads in teacher (Phases~1--2) and student (Phase~3).}
\label{tab:dropout_schedules}

\begin{tabular}{@{}lcc@{}}
\toprule
 & \textbf{Phases 1--2 (Teacher)} & \textbf{Phase 3 (Student)} \\
\midrule
Projection head $(p_1, p_2)$ & $(0.5,\ 0.4)$ & $(0.3,\ 0.2)$ \\
Classification head $p$      & $0.6$         & $0.4$ \\
\bottomrule
\end{tabular}
\end{table}

\paragraph{Phase 1 - Complete Hyperparameters}
\label{suppl:phase1_hyperparameters}
\begin{itemize}
    \item Epochs: 30
    \item Batch size: 6
    \item Optimizer: AdamW with $\text{lr}=2 \times 10^{-5}$, $\text{wd}=1 \times 10^{-3}$ (uniform)
    \item Slices per subject: 25 (uniformly spaced)
    \item Loss: BCE (no label smoothing)
    \item Gradient clipping: max\_norm dynamically adjusted (1.0 for epochs 1--2, 2.0 for epochs 3--5, 5.0 thereafter)
\end{itemize}

\paragraph{Phase 2 - Complete Hyperparameters}
\label{suppl:phase2_hyperparameters}
\begin{itemize}
    \item Epochs: 15
    \item Batch size: 6
    \item Optimizer: AdamW with component-specific rates:
    \begin{itemize}
        \item Vision backbone, projection, attention: $\text{lr}=5 \times 10^{-6}$, $\text{wd}=10^{-2}$
        \item Classification head: $\text{lr}=2 \times 10^{-5}$, $\text{wd}=5 \times 10^{-3}$
    \end{itemize}
    \item Scheduler: CosineAnnealingWarmRestarts with $T_0 = 5$, $T_{\text{mult}} = 2$, $\eta_{\min} = 10^{-7}$
\end{itemize}

\paragraph{Phase 2 Regularization}
\label{suppl:phase2_reg}
$\mathcal{L}_{\text{reg}}$ includes three components: (i)~$\ell_2$ penalty on anchor, positive, and negative embedding norms with coefficient 0.01; (ii)~inter-anchor similarity penalty, penalizing mean pairwise cosine similarity above 0.5; (iii)~anchor-negative similarity penalty (weight 0.5), penalizing mean similarity above $-0.1$. Components (ii--iii) use progressive epoch-dependent scaling: $s = 0.1$ for epochs 1--3, then linearly increasing to 1.0 by epoch~13.

\paragraph{MarginFocal Loss Details (Phase 3)}
\label{suppl:marginfocal}
Complete hyperparameters: $\gamma = 2.0$ (focal parameter), $w = 1.0$ (pos\_weight for balanced batches), label smoothing $= 0$, $\varepsilon = 10^{-8}$ (numerical stability floor).

The positive-weighted BCE is: $\mathrm{BCE}_w(\tilde z, y') = -w y'\log\sigma(\tilde z) - (1-y')\log(1-\sigma(\tilde z))$.

Margin annealing schedule:
\begin{itemize}
    \item Epochs 1--6: $m = 0.3$
    \item Epochs 7--20: $m$ is linearly increased from $0.3$ towards the final value
    \item Epochs 21+: $m = 1.2$ (fixed)
\end{itemize}

Gap deficit scaling: The term $[\,m - (\bar z_{+} - \bar z_{-})\,]_+$ is multiplied by $0.1$ internally before adding to the loss, yielding effective $\lambda_{\mathrm{gap}} \approx 0.01$ (epochs 1--10) and $0.03$ (epochs 11+).

\paragraph{Phase 3 - Complete Hyperparameters}
\label{suppl:phase3_hyperparameters}
\begin{itemize}
    \item Epochs: 100
    \item Batch size: 10
    \item Optimizer: AdamW with component-specific rates:
    \begin{itemize}
        \item LoRA adapters: $\text{lr}=2 \times 10^{-4}$, $\text{wd} = 0$ (preserve low-rank structure)
        \item Projection modules: $\text{lr}=1 \times 10^{-4}$, $\text{wd}=1 \times 10^{-4}$
        \item Attention modules: $\text{lr}=1 \times 10^{-4}$, $\text{wd}=1 \times 10^{-3}$
        \item Classification head: $\text{lr}=1 \times 10^{-4}$, $\text{wd}=1 \times 10^{-3}$
    \end{itemize}
    \item Scheduler: ReduceLROnPlateau (mode=max, factor=0.7, patience=15, threshold=0.005, $\texttt{min\_lr} = 10^{-5}$, cooldown=2; monitoring validation F1)
    \item Temperature annealing: $T = 2.5$ (epochs 1--6), linearly decreased (epochs 7--20) towards $T = 1.0$, and fixed at $T = 1.0$ (epochs 21+)
\end{itemize}

\paragraph{Knowledge Distillation Loss Weights Warm-up}
During Phase~3, we linearly warm up the loss weights over the first 10 epochs from $(\lambda_{\text{cls}}, \lambda_{\text{feat}}, \lambda_{\text{logit}}) = (0.3, 0.5, 0.2)$ to $(0.4, 0.4, 0.2)$. After epoch~10, the weights are kept fixed at $(\lambda_{\text{cls}}, \lambda_{\text{feat}}, \lambda_{\text{logit}}) = (0.4, 0.4, 0.2)$.

\section{Additional Experimental Results}
\label{suppl:additional_experiments}
\subsection{Multi-Contrast Performance Analysis}

\begin{figure}[!htb]
  \centering
  \begin{subfigure}[b]{0.49\columnwidth}
    \centering
    \includegraphics[width=\textwidth]{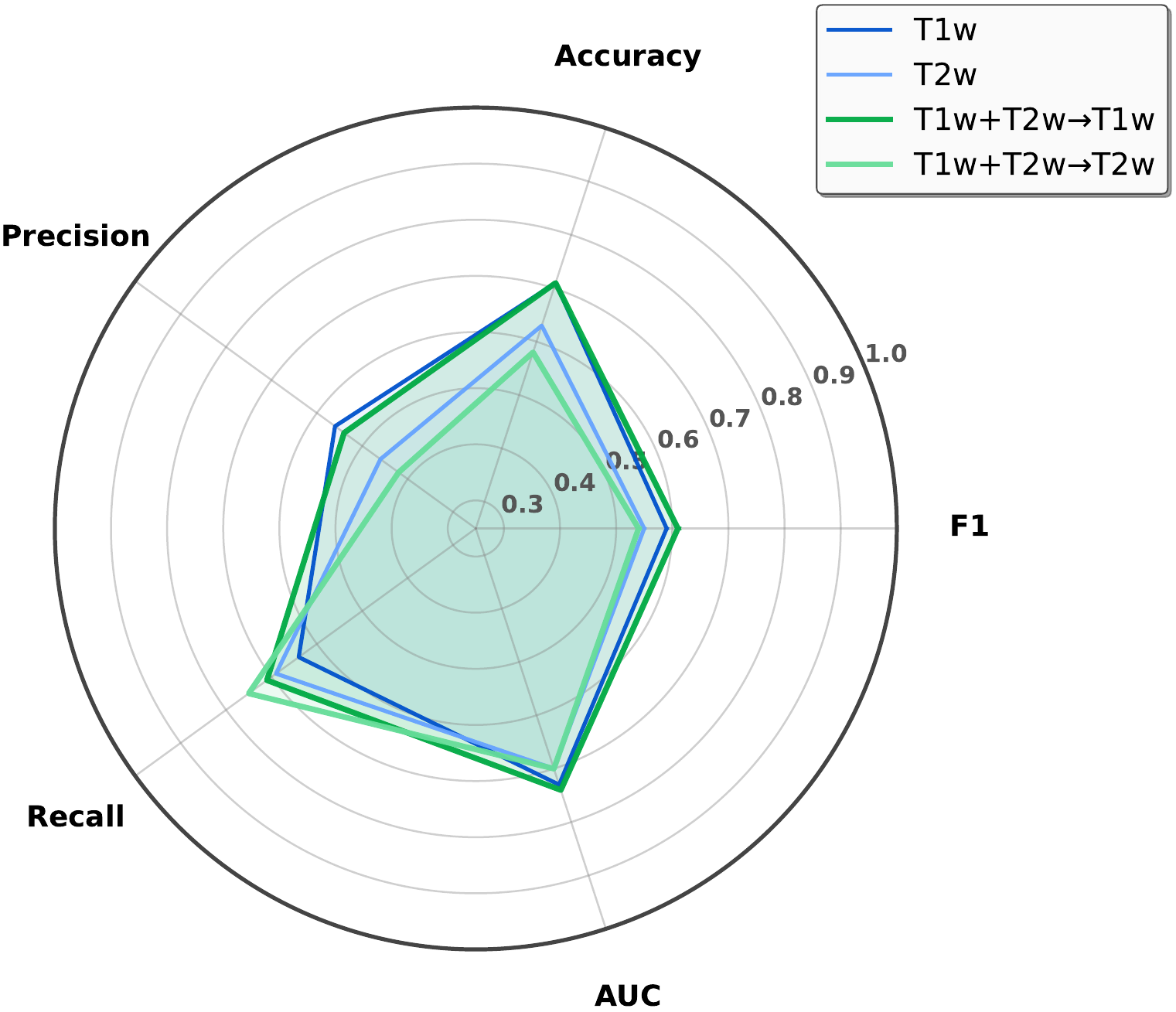}
    \caption{OASIS-3: T1w \& T2w}
  \end{subfigure}\hfill
  \begin{subfigure}[b]{0.49\columnwidth}
    \centering
    \includegraphics[width=\textwidth]{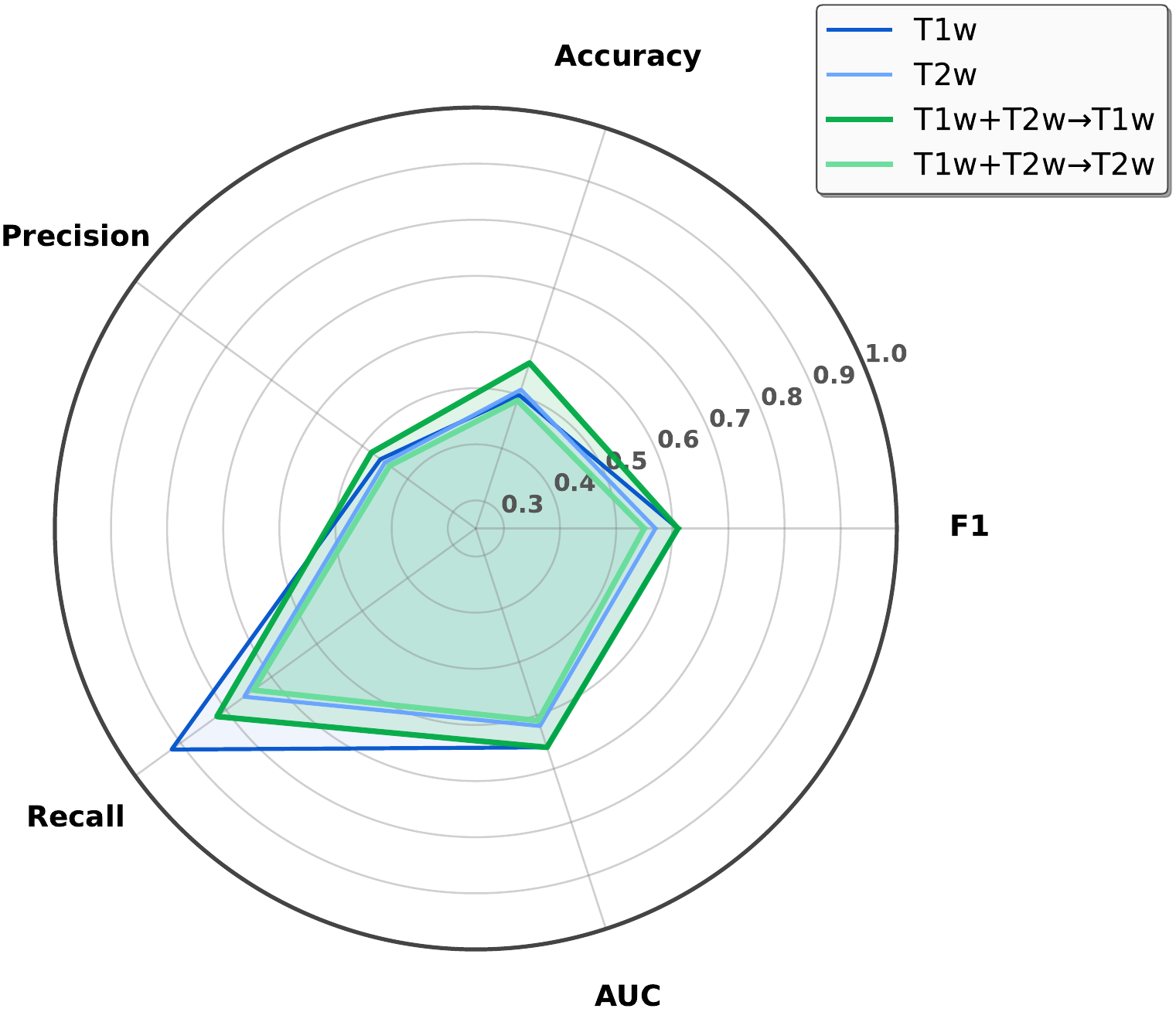}
    \caption{ADNI: T1w \& T2w}
  \end{subfigure}
  
  \vspace{0.3cm}
  
  \begin{subfigure}[b]{0.49\columnwidth}
    \centering
    \includegraphics[width=\textwidth]{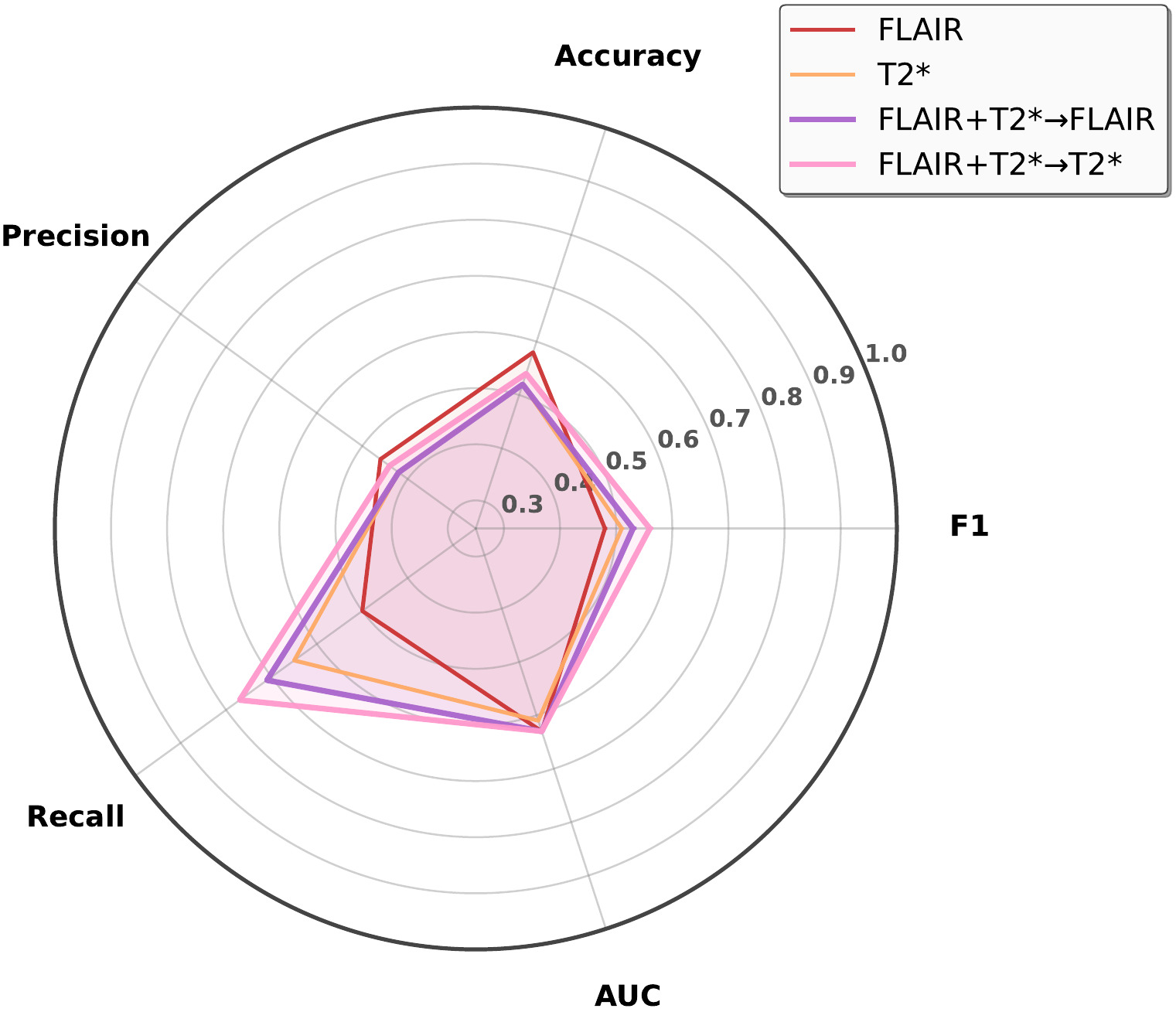}
    \caption{OASIS-3: FLAIR \& T2*}
  \end{subfigure}\hfill
  \begin{subfigure}[b]{0.49\columnwidth}
    \centering
    \includegraphics[width=\textwidth]{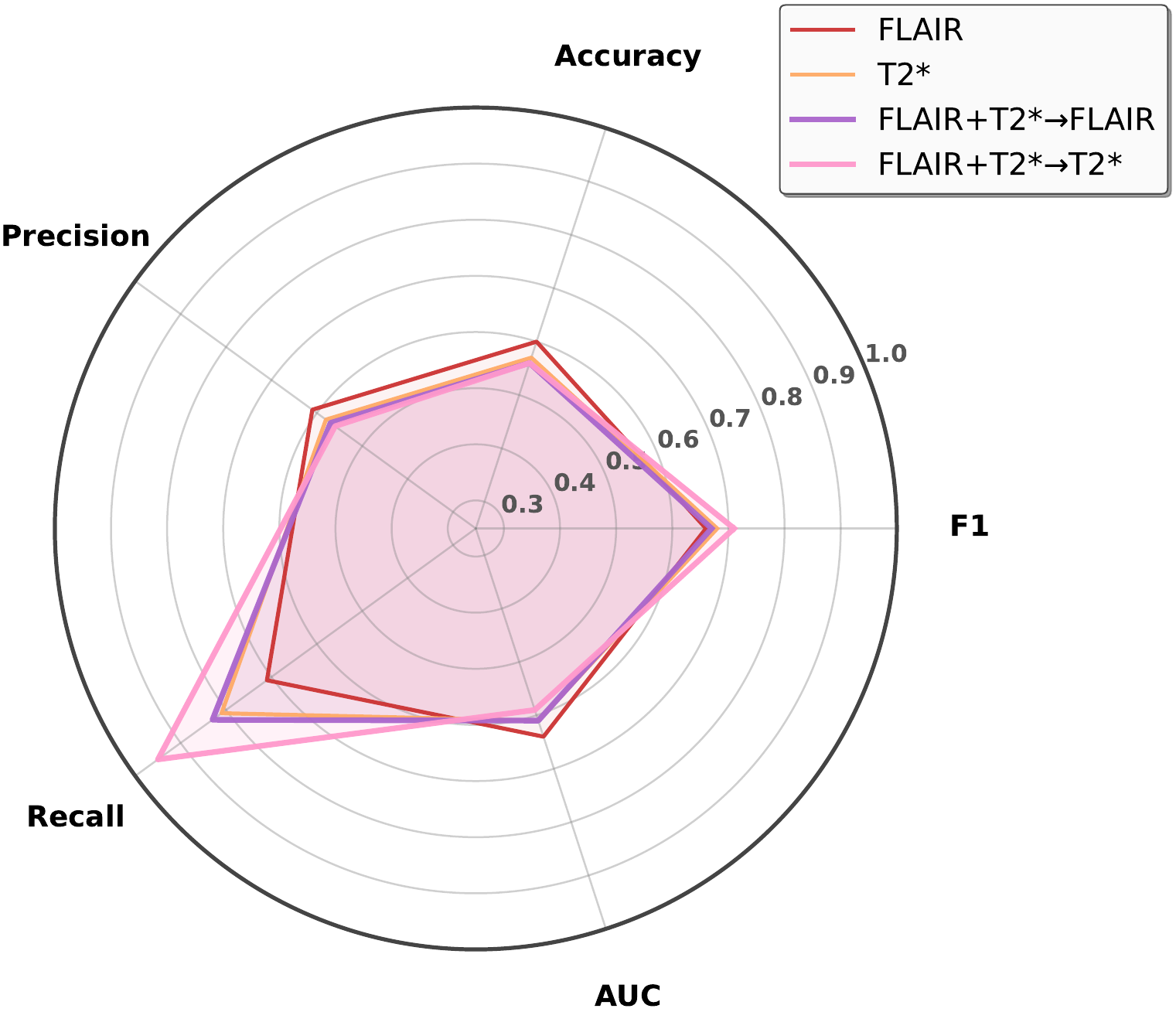}
    \caption{ADNI: FLAIR \& T2*}
  \end{subfigure}
  
    \caption{\textbf{Single vs Multi-Sequence Performance Comparison.} 
    Spider charts comparing single-sequence and multi-sequence distilled models across metrics (F1, Accuracy, Precision, Recall, AUC). Top: T1w+T2w; bottom: FLAIR+T2*. Left: OASIS-3; right: ADNI. Multi-sequence distillation improves recall.}
  \label{fig:radar_comparison}
\end{figure}

Fig.~\ref{fig:radar_comparison} compares models that use a single sequence with models that use multiple sequences and are then tested on a single sequence using five metrics (F1, Accuracy, Precision, Recall, AUC). Multi-sequence models consistently improve recall (e.g., T1w on OASIS-3: 0.64 $\to$ 0.71, +10.9\%), while also increasing AUC on OASIS-3 (0.73 $\to$ 0.74) and accuracy for T1w on ADNI (0.50 $\to$ 0.56), and maintaining comparable precision and AUC in both cohorts. On OASIS-3 (Fig.~\ref{fig:radar_comparison}c), FLAIR+T2* $\to$ T2* achieves the largest F1 improvement (0.51 $\to$ 0.56, +9.8\%) with substantial recall gains (0.65 $\to$ 0.77). In ADNI, the recall gain is larger; in particular, FLAIR+T2* $\to$ T2* reaches F1~0.71 with recall~0.95. Across all four charts, multi-sequence distillation yields higher recall while maintaining or improving the other metrics, suggesting that using multiple sequences provides the model with richer context when predicting from a single sequence and is therefore preferable in settings where minimizing false negatives is critical.

\subsection{Ablation Study: Impact of Centiloid-Based Negative Mining}
\label{suppl:ablation_centiloid}

\begin{table*}[!htb]
\centering
\caption{\textbf{Ablation study: impact of Centiloid-based negative mining.} Test performance on OASIS-3 and ADNI for T1w and T2w models when varying the minimum Centiloid gap $\Delta_{\text{CL}}^{\min}$ between anchor and negative samples. We report metrics both at a fixed decision threshold of $0.5$ and at the validation-optimized threshold $\theta^*$.}

\label{tab:ablation_centiloid}
\setlength{\tabcolsep}{2.6pt}
\renewcommand{\arraystretch}{1.05}
{\scriptsize
\resizebox{\linewidth}{!}{%
\begin{tabular}{@{}l lcccccccccc lcccccccccc@{}}
\toprule
& \multicolumn{11}{c}{\textbf{OASIS-3}} & \multicolumn{11}{c}{\textbf{ADNI}} \\
\cmidrule(lr){2-12}\cmidrule(l){13-23}
\multirow{2}{*}{Sequence}
  & \multirow{2}{*}{$\Delta_{\text{CL}}^{\min}$}
  & \multicolumn{4}{c}{@0.5}
  & \multicolumn{4}{c}{@$\theta^*$}
  & AUC & NPV
  & \multirow{2}{*}{$\Delta_{\text{CL}}^{\min}$}
  & \multicolumn{4}{c}{@0.5}
  & \multicolumn{4}{c}{@$\theta^*$}
  & AUC & NPV \\
& & F1 & Acc & Prec & Rec & F1 & Acc & Prec & Rec & & &
    & F1 & Acc & Prec & Rec & F1 & Acc & Prec & Rec & & \\
\midrule
\multicolumn{23}{@{}l}{\textbf{T1-weighted MRI}} \\
T1w & 0.0   & 0.53 & 0.49 & 0.38 & 0.87 & 0.53 & 0.68 & 0.51 & 0.56 & 0.72 & 0.77
    & 0.0   & 0.60 & 0.63 & 0.56 & 0.66 & 0.60 & 0.60 & 0.52 & 0.71 & 0.65 & 0.71 \\
T1w & 5.0   & 0.53 & 0.71 & 0.57 & 0.49 & 0.59 & 0.71 & 0.56 & 0.64 & 0.73 & 0.81
    & 5.0   & 0.58 & 0.56 & 0.48 & 0.74 & 0.61 & 0.50 & 0.46 & 0.92 & 0.66 & 0.77 \\
T1w & 10.0  & 0.53 & 0.65 & 0.47 & 0.60 & 0.52 & 0.63 & 0.46 & 0.60 & 0.68 & 0.77
    & 10.0  & 0.59 & 0.60 & 0.52 & 0.68 & 0.58 & 0.53 & 0.47 & 0.76 & 0.65 & 0.68 \\
\midrule
\multicolumn{23}{@{}l}{\textbf{T2-weighted MRI}} \\
T2w & 0.0   & 0.47 & 0.66 & 0.49 & 0.46 & 0.55 & 0.60 & 0.44 & 0.75 & 0.69 & 0.81
    & 0.0   & 0.55 & 0.53 & 0.46 & 0.68 & 0.56 & 0.51 & 0.45 & 0.74 & 0.59 & 0.64 \\
T2w & 5.0   & 0.54 & 0.54 & 0.40 & 0.80 & 0.55 & 0.63 & 0.46 & 0.69 & 0.70 & 0.80
    & 5.0   & 0.51 & 0.63 & 0.59 & 0.45 & 0.57 & 0.51 & 0.45 & 0.76 & 0.62 & 0.65 \\
T2w & 10.0  & 0.55 & 0.61 & 0.44 & 0.71 & 0.55 & 0.61 & 0.44 & 0.71 & 0.70 & 0.80
    & 10.0  & 0.52 & 0.59 & 0.51 & 0.53 & 0.58 & 0.51 & 0.46 & 0.79 & 0.60 & 0.67 \\
\bottomrule
\end{tabular}
}
}
\end{table*}

To assess the contribution of the Centiloid-guided negative mining strategy used in Phase~2, we vary the minimum amyloid burden difference threshold $\Delta_{\text{CL}}^{\min}$ required between anchor and negative samples (Table~\ref{tab:ablation_centiloid}) and evaluate three configurations for both T1w and T2w sequences:
\begin{itemize}
    \item \textbf{Uniform ($\Delta_{\text{CL}}^{\min} = 0$):} Negatives sampled uniformly, representing a baseline where triplet learning relies on visual similarity.
    \item \textbf{Moderate ($\Delta_{\text{CL}}^{\min} = 5.0$):}
    configuration used in the main experiments, requiring negatives to differ by at least 5~CL units from the anchor.
    \item \textbf{Strict ($\Delta_{\text{CL}}^{\min} = 10.0$):}
    more restrictive threshold enforcing larger amyloid burden differences, yielding harder negatives but reducing the number of eligible triplets.
\end{itemize}

\paragraph{Analysis of Results}
Table~\ref{tab:ablation_centiloid} shows that using a specific Centiloid gap constraint value ($\Delta_{\text{CL}}^{\min} > 0$) is beneficial compared to using a uniformly selected negative patient. The choice $\Delta_{\text{CL}}^{\min}=5.0$ provides the best trade-off between sensitivity and specificity across datasets, with better calibration of the metrics. For T1w on OASIS-3, the threshold of 5.0 improves the F1 score from 0.53 to 0.59 (+11.3\%) and increases NPV from 0.77 to 0.81 compared to ($\Delta_{\text{CL}}^{\min}=0$), reducing the number of missed amyloid-positive cases. The recall improvement (0.56$\to$0.64) implies a relative reduction of approximately 18\% in the number of false negatives in the test set. The \textbf{Strict} strategy ($\Delta_{\text{CL}}^{\min} = 10.0$) underperforms on OASIS-3 T1w (F1 = 0.52, AUC = 0.68), suggesting that using too high a Centiloid value prevents the network from learning smaller differences between patients that are not too dissimilar, as obtained with $\Delta_{\text{CL}}^{\min}=5.0$. For T2w on OASIS-3, all three strategies lead to similar F1 scores, but using a moderate $\Delta_{\text{CL}}^{\min}$ achieves an AUC of 0.70 (similar to the strict configuration and higher than the uniform baseline at 0.69), indicating better calibration.

On ADNI, the trends are similar but less pronounced. For T1w, using the Moderate configuration (F1 = 0.61, AUC = 0.66) marginally outperforms both Uniform (F1 = 0.60, AUC = 0.65) and Strict (F1 = 0.58, AUC = 0.65). A similar pattern is observed for T2w, with Moderate achieving F1 = 0.57 and AUC = 0.62. We also note that the Uniform baseline remains competitive on ADNI (e.g., T1w: F1 = 0.60), likely due to its more balanced class distribution, which reduces the risk of trivial negatives. The smaller performance gap on ADNI suggests that CL-aware mining provides greater benefits in imbalanced settings. These results validate our choice of $\Delta_{\text{CL}}^{\min} = 5.0$ as the optimal balance: it enforces biochemically meaningful separation by requiring negatives to differ by at least 5~Centiloid units from the anchor, a threshold that exceeds both the test-retest measurement error (2.5--3.5~CL) and the reliable annual amyloid accumulation rate (3--5~CL/year)~\cite{collijCentiloidRecommendationsClinical2024}, while maintaining sufficient triplet diversity for effective contrastive learning. Across both datasets and contrasts, the Moderate setting either matches or outperforms the Uniform and Strict strategies, with the largest absolute gains observed on OASIS-3.

\subsection{Training Convergence Analysis}
\label{suppl:roc_analysis}

\begin{figure*}[p]
  \centering
  \begin{subfigure}[b]{0.49\textwidth}
    \includegraphics[width=\textwidth]{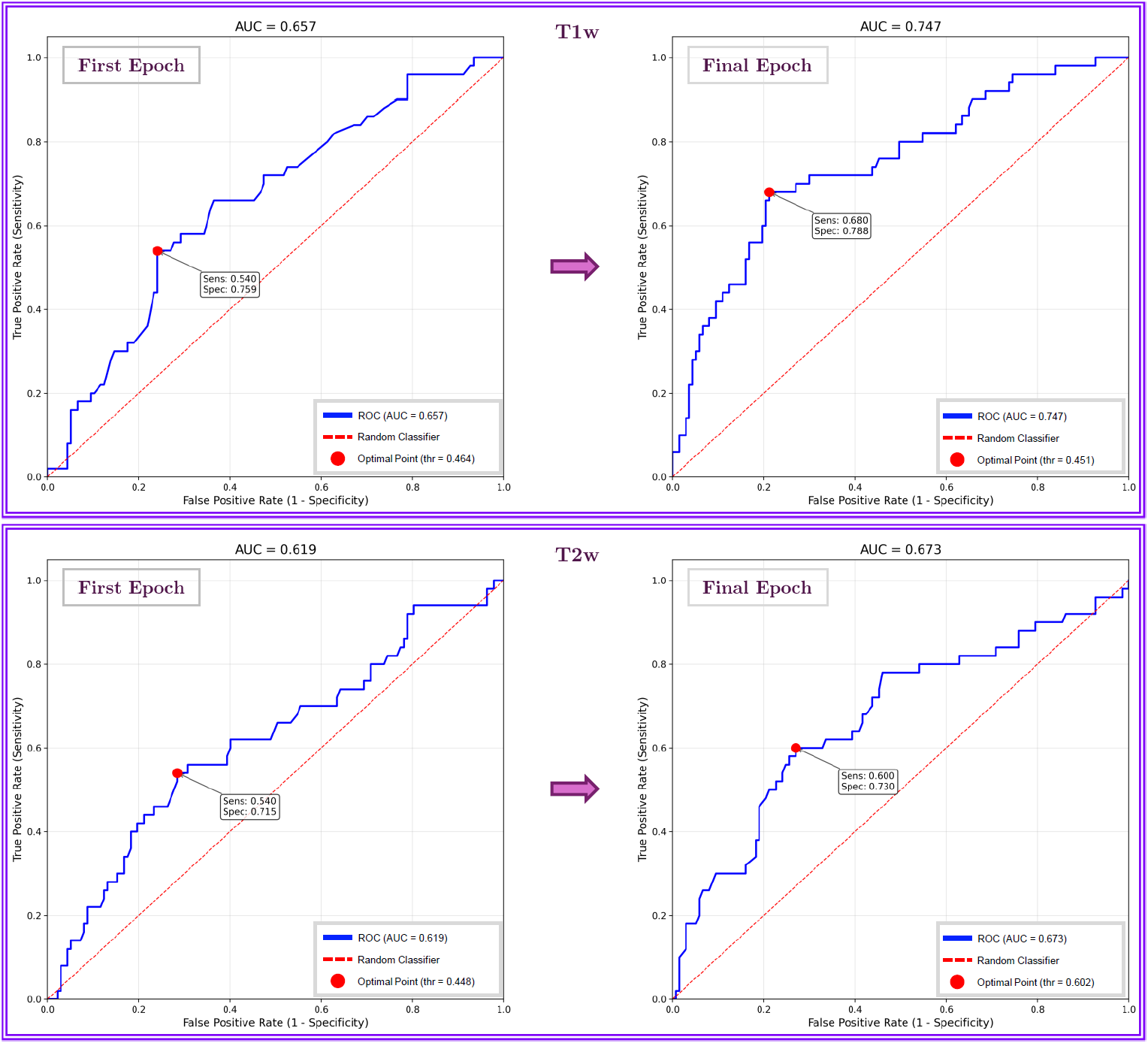}
    \caption{OASIS-3: T1w and T2w}
    \label{fig:roc_oasis_t1t2}
  \end{subfigure}\hfill
  \begin{subfigure}[b]{0.49\textwidth}
    \includegraphics[width=\textwidth]{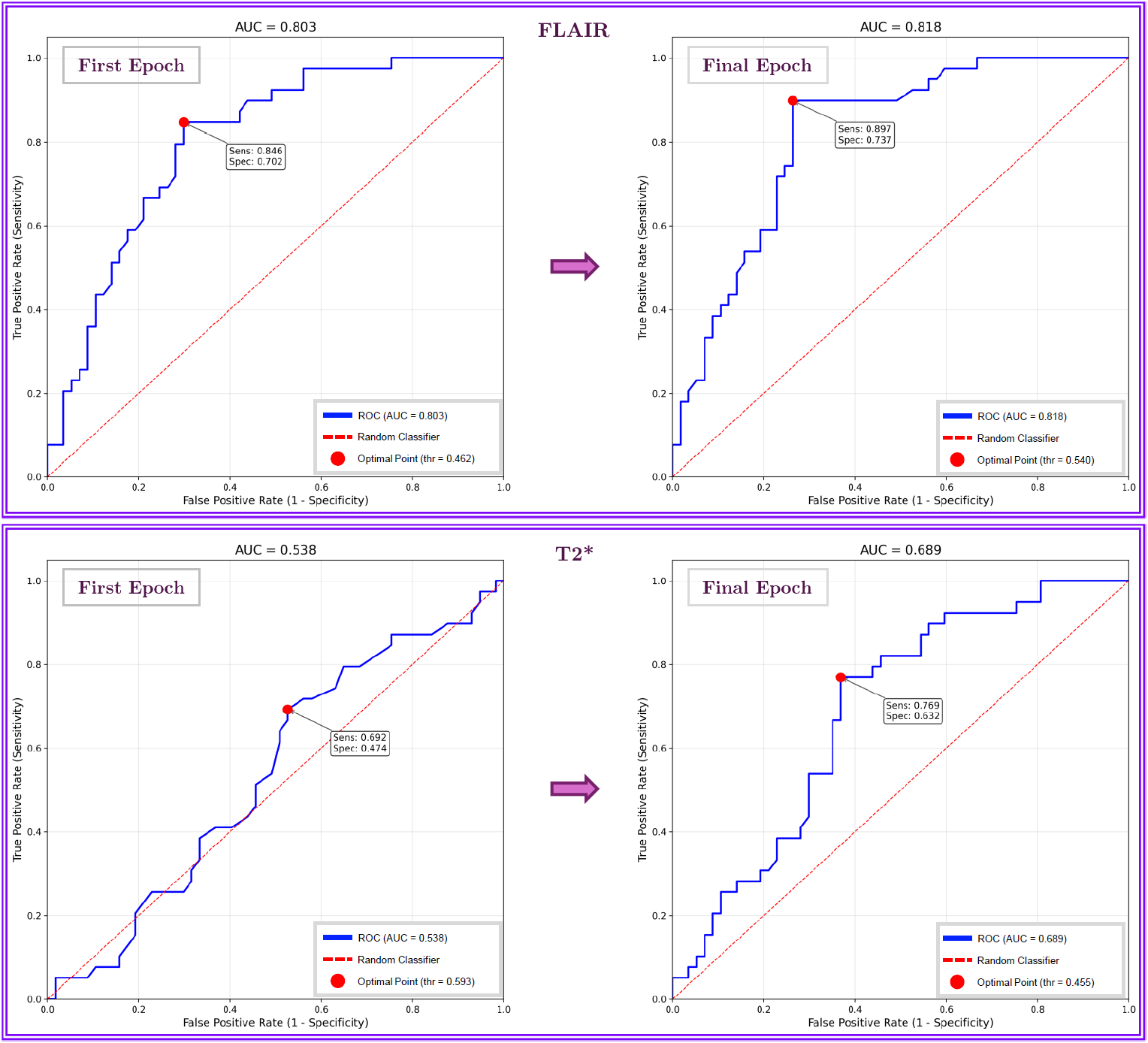}
    \caption{OASIS-3: FLAIR and T2*}
    \label{fig:roc_oasis_flair}
  \end{subfigure}
  
  \vspace{0.2cm}
  
  \begin{subfigure}[b]{0.49\textwidth}
    \includegraphics[width=\textwidth]{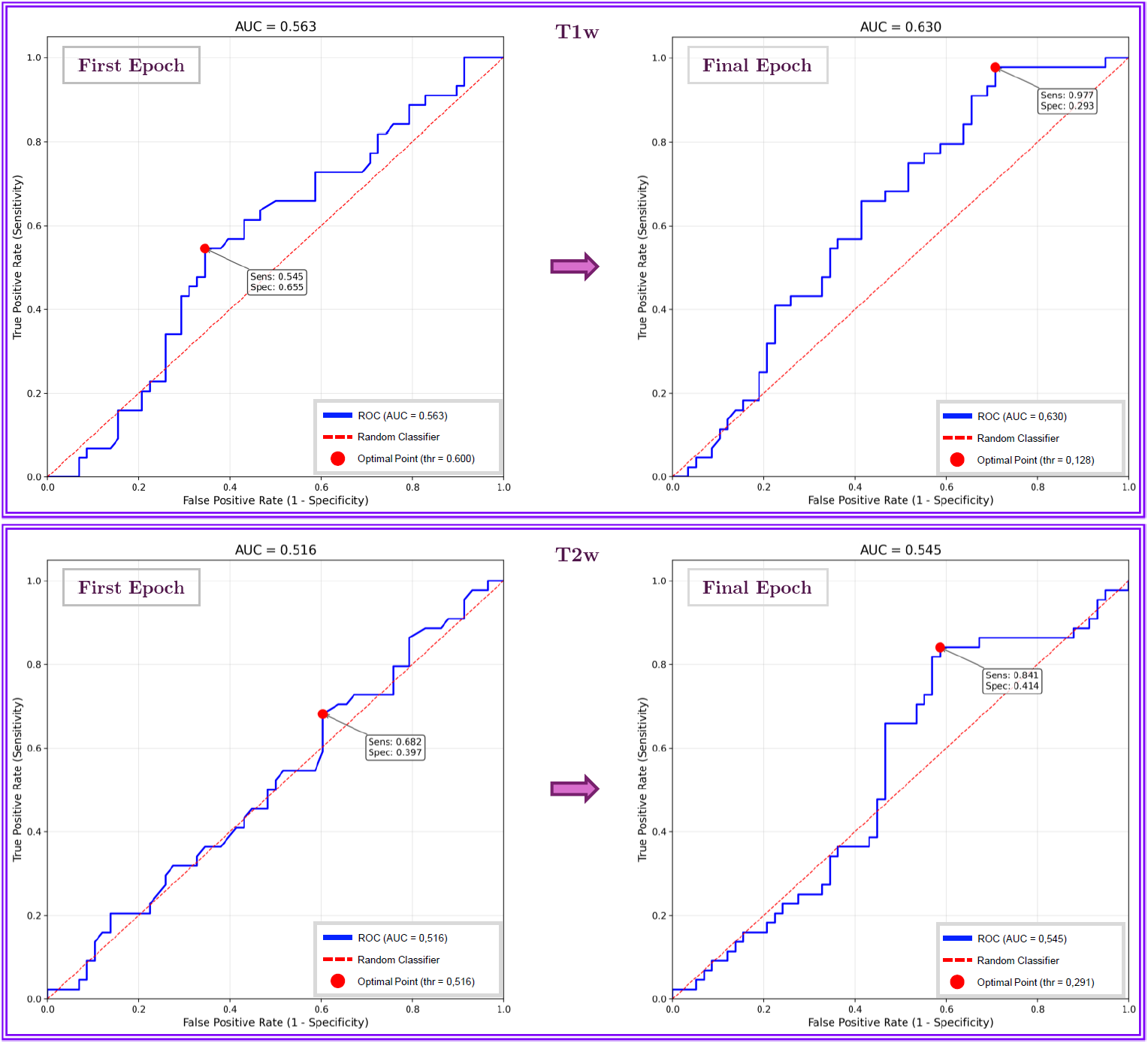}
    \caption{ADNI: T1w and T2w}
    \label{fig:roc_adni_t1t2}
  \end{subfigure}\hfill
  \begin{subfigure}[b]{0.49\textwidth}
    \includegraphics[width=\textwidth]{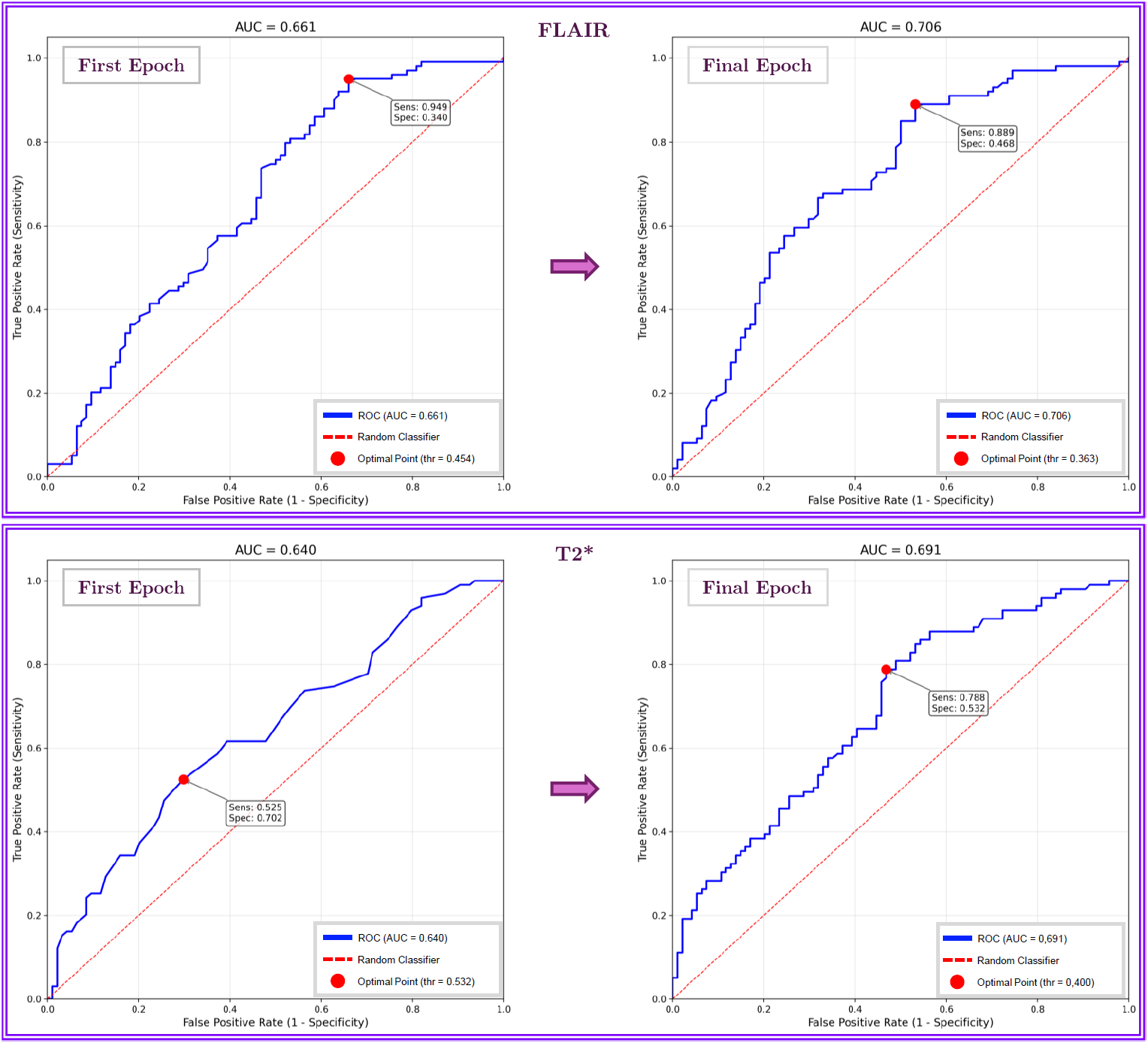}
    \caption{ADNI: FLAIR and T2*}
    \label{fig:roc_adni_flair}
  \end{subfigure}
  
  \caption{\textbf{ROC curve evolution across datasets and sequences.} 
  Validation performance at Epoch 1 (initialization) vs. Final Epoch (convergence). 
  Top row: OASIS-3 dataset for T1w+T2w training (left) and FLAIR+T2* training (right). Bottom row: ADNI dataset with same training configurations. All models show substantial AUC improvements demonstrating effective knowledge distillation.}
  \label{fig:roc_combined}
\end{figure*}

Fig.~\ref{fig:roc_combined} visualizes the evolution of the ROC curves from the first epoch (Epoch 1) to the last epoch (Final Epoch) on the validation set. We plot the final training epoch rather than the early-stopping checkpoint. All models show positive $\Delta$AUC improvements from the first to the final epoch, confirming effective knowledge distillation. OASIS-3 achieves superior results in terms of AUCs compared to ADNI, possibly due to differences in cohort composition, image acquisition protocols, or amyloid distribution across datasets~\cite{jackOverviewADNIMRI2024,samper-gonzalezReproducibleEvaluationClassification2018}.

\subsection{Interpretability}
\label{suppl:interpretability}
\begin{figure*}[!htb]
  \centering
    \begin{subfigure}{\textwidth}
    \includegraphics[width=\textwidth]{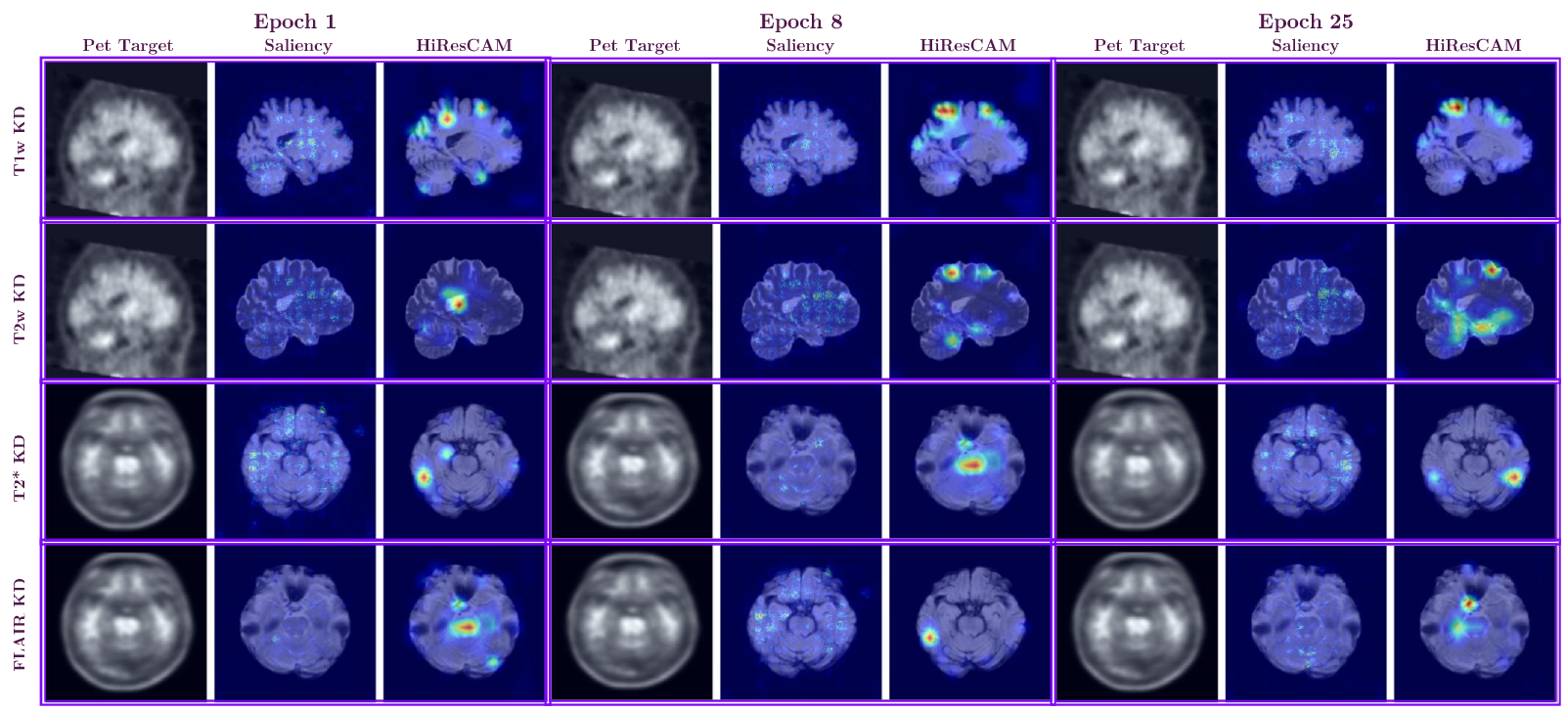}
    \caption{OASIS-3 dataset}
    \label{fig:suppl_interp_oasis}
  \end{subfigure}
  
  \vspace{0.2cm}
  
    \begin{subfigure}{\textwidth}
    \includegraphics[width=\textwidth]{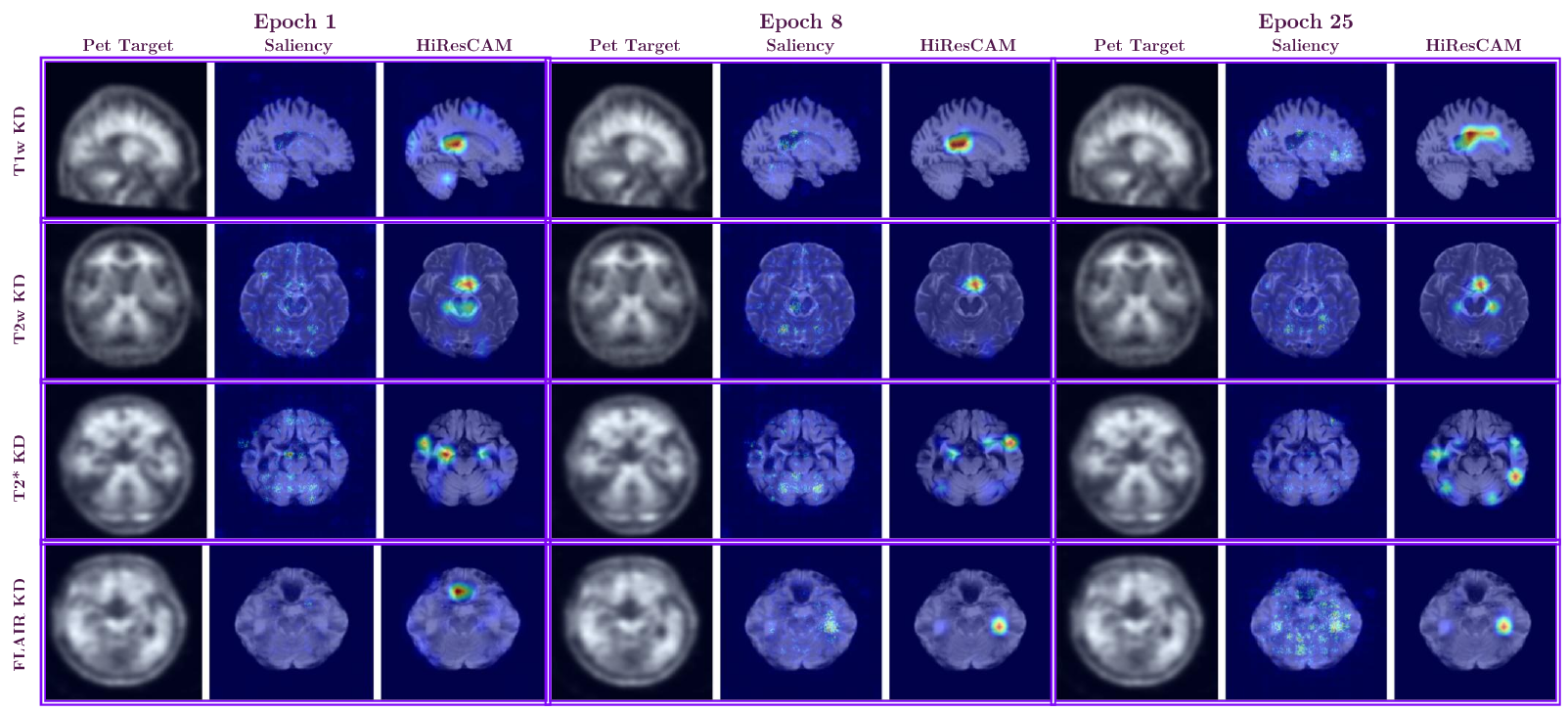}
    \caption{ADNI dataset}
    \label{fig:suppl_interp_adni}
  \end{subfigure}
  
    \caption{\textbf{Single-sequence training: attention evolution across datasets.} 
    Training progression (epochs 1, 8, 25) for models trained on individual MRI contrasts. Each row displays PET reference, target MRI, gradient saliency, and HiResCAM maps. The network progressively focuses on anatomically relevant brain structures, with consistent patterns across OASIS-3 and ADNI datasets demonstrating robust generalization.}
  \label{fig:suppl_interp_combined}
\end{figure*}

\begin{figure*}[!htb]
  \centering
    \begin{subfigure}{\textwidth}
    \includegraphics[width=\textwidth]{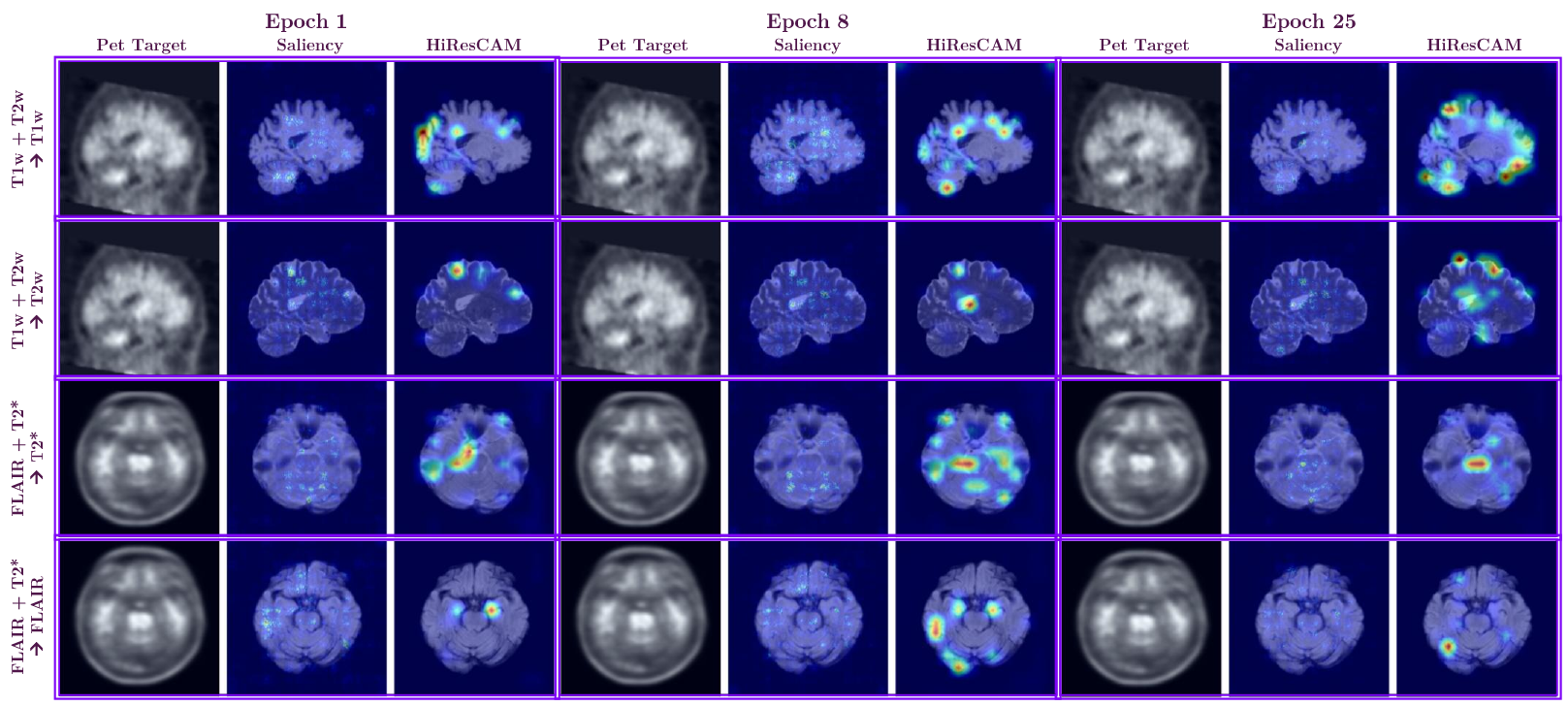}
    \caption{OASIS-3 dataset}
    \label{fig:suppl_interp_multi_oasis}
  \end{subfigure}
  
  \vspace{0.2cm}
  
    \begin{subfigure}{\textwidth}
    \includegraphics[width=\textwidth]{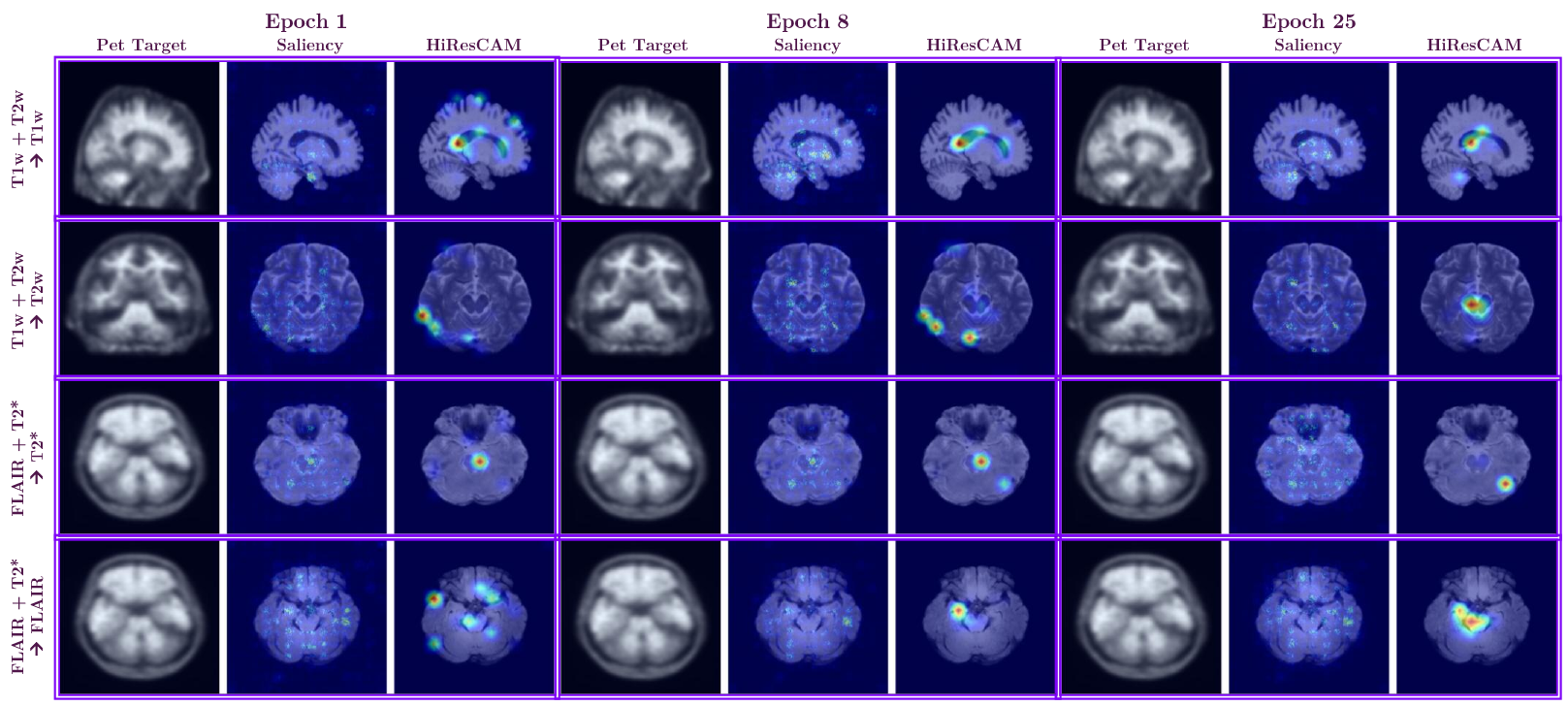}
    \caption{ADNI dataset}
    \label{fig:suppl_interp_multi_adni}
  \end{subfigure}
  
    \caption{\textbf{Multi-sequence training with single-sequence inference.} 
    Saliency/HiResCAM evolution (epochs 1, 8, 25) for models trained on paired 
    sequences (T1w+T2w, FLAIR+T2*) and tested on individual contrasts, showing 
    consistent spatial attention patterns across modalities.}
  \label{fig:suppl_interp_multi_combined}
\end{figure*}

Figs.~\ref{fig:suppl_interp_combined} and~\ref{fig:suppl_interp_multi_combined} show how the model’s spatial attention evolves during training. We visualize three epochs (1, 8, and 25) for both
single-sequence models (T1w, T2w, FLAIR, T2*) and multi-sequence models (T1w+T2w and FLAIR+T2*),
each tested on an individual contrast. For each dataset (OASIS-3 and ADNI) and configuration, we display the target PET image with gradient-based saliency maps and HiResCAM explanations. At epoch~1, both saliency and HiResCAM are relatively diffuse across the brain volume, indicating that the networks initially rely on non-specific global patterns. By epoch~8, the model's attention becomes more structured and begins to concentrate on regions that more closely correspond to the reference PET signal. By epoch~25, the maps are more focal, highlighting neuroanatomically plausible regions. Qualitatively, the attention patterns are consistent between OASIS-3 and ADNI, suggesting that the learned features capture generalizable amyloid-related patterns rather than dataset-specific artifacts.

\FloatBarrier

\end{document}